\documentclass[10pt,twocolumn,letterpaper]{article}
\pdfoutput=1

\usepackage[pagenumbers]{cvpr} 
\usepackage[accsupp]{axessibility}

\usepackage{graphicx}
\usepackage{amsmath}
\usepackage{amssymb}
\usepackage{booktabs}
\usepackage{xcolor}
\usepackage{overpic}
\newcommand{\generator}{\ensuremath{\mathcal{G}}} 
\newcommand{\bbox}{\mathcal{B}} 
\newcommand{\border}{\mathcal{E}} 
\newcommand{\downsample}{\mathcal{D}} 
\newcommand{\percep}{L_{\text{\textit{lpips}}}} 
\newcommand{\Lone}{L_1} 
\newcommand{\Ltwo}{L_2} 
\renewcommand{\vec}[1]{\mathbf{#1}}
\newcommand{\gA}{\generator_A} 
\newcommand{\gB}{\generator_B} 
\newcommand{\w}{\vec{w}} 
\newcommand{\wA}{\w_A} 
\newcommand{\wB}{\w_B} 
\newcommand{\iA}{I_A} 
\newcommand{\iB}{I_B} 
\newcommand{\dA}{I^\downarrow_A} 
\newcommand{\dB}{I^\downarrow_B} 
\newcommand{\lA}{L_{\text{\textit{coarse}}}} 
\newcommand{\lB}{L_{\text{\textit{border}}}} 
\newcommand{\lRB}{L_{\text{\textit{body}}}} 
\newcommand{\lRF}{L_{\text{\textit{face}}}} 
\newcommand{\lR}{L_{\text{\textit{reg}}}} 
\newcommand{\region}{R} 
\newcommand{\weight}{\lambda} 
\newcommand{\iRef}{I_{\text{\textit{ref}}}} 
\newcommand{\wAvg}{\w_{\text{\textit{avg}}}} 
\newcommand{\wTrunc}{\w_{\text{\textit{trunc}}}} 
\newcommand{\wRand}{\w_{\text{\textit{rand}}}} 

\usepackage[pagebackref,breaklinks,colorlinks]{hyperref}

\usepackage[capitalize]{cleveref}
\crefname{section}{Sec.}{Secs.}
\Crefname{section}{Section}{Sections}
\Crefname{table}{Table}{Tables}
\crefname{table}{Tab.}{Tabs.}

\newcommand{\name}{InsetGAN\xspace}
\newcommand{\comodgan}{CoModGAN\xspace}

\def\cvprPaperID{9245}
\def\confName{CVPR}
\def\confYear{2022}

\begin{document}

\title{\name for Full-Body Image Generation\vspace*{-.15in}}
\author{
    \begin{tabular}{c} Anna Fr\"{u}hst\"{u}ck\textsuperscript{{1, 2}} \\ Niloy J. Mitra\textsuperscript{{2, 3}} \end{tabular}
    \begin{tabular}{c} Krishna Kumar Singh\textsuperscript{2} \\ Peter Wonka\textsuperscript{1} \end{tabular}
    \begin{tabular}{c} Eli Shechtman\textsuperscript{2} \\ Jingwan Lu\textsuperscript{2} \end{tabular}\\[2ex]
    \textsuperscript{1~}KAUST \quad 
	\textsuperscript{2~}Adobe Research \quad
	\textsuperscript{3~}University College London\\[1.1ex]
	{\tt\footnotesize anna.fruehstueck@kaust.edu.sa, \{krishsin, elishe\}@adobe.com, \{niloym, pwonka\}@gmail.com, jlu@adobe.com}\\
}

\newcommand\blfootnote[1]{%
  \begingroup
  \renewcommand\thefootnote{}\footnote{#1}%
  \addtocounter{footnote}{-1}%
  \endgroup
}

\twocolumn[{%
\renewcommand\twocolumn[1][]{#1}%
\maketitle
\vspace*{-.35in}
\begin{center}
  \centering
  \captionsetup{type=figure}
  \includegraphics[width=0.99\textwidth]{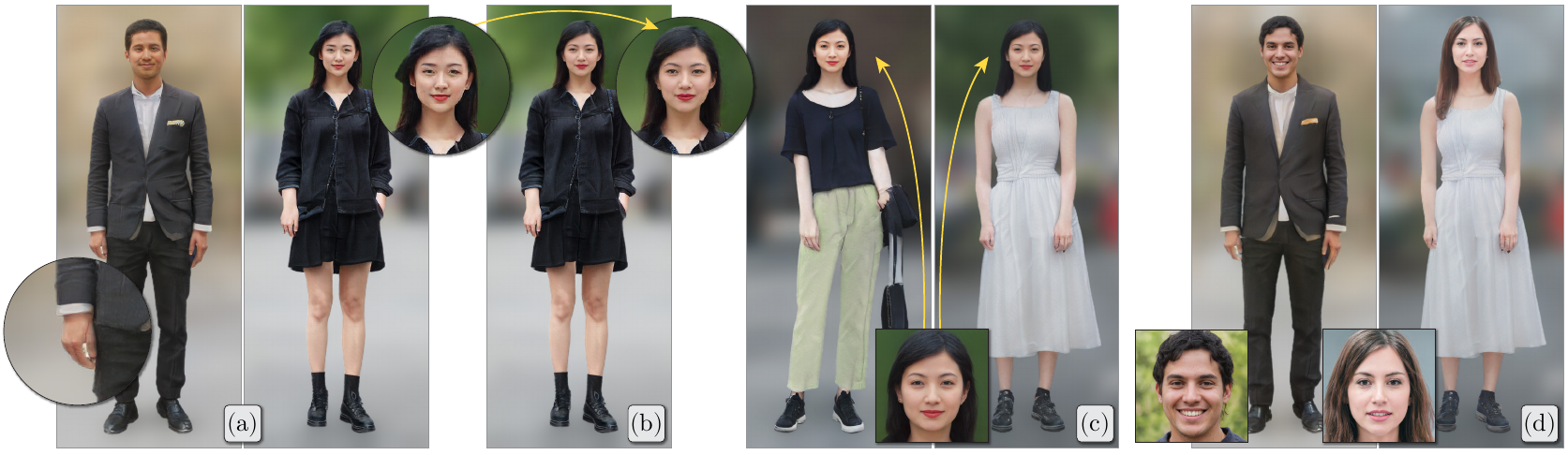}
  \vspace*{-.07in}
  \caption{\textbf{\name applications.} Our full-body human generator is able to generate reasonable bodies at state-of-the-art resolution (1024$\times$1024px) \textbf{(a)}. However, some artifacts appear in the synthesized results, most visibly in extremities and faces. We make use of a second, specialized generator to seamlessly improve the face region \textbf{(b)}. We can also use a given face as an input for unconditional generation of bodies \textbf{(c)}. Furthermore, we can select both specific faces and bodies and compose them in a seamlessly merged output \textbf{(d)}. } 
  \label{fig:teaser}
\end{center}%
}]

\begin{abstract}

While GANs can produce photo-realistic images in ideal conditions for certain domains, the generation of full-body human images remains difficult due to the diversity of identities, hairstyles, clothing, and the variance in pose.
Instead of modeling this complex domain with a single GAN, we propose a novel method to combine multiple pretrained GANs, where one GAN generates a global canvas (e.g., human body) and a set of specialized GANs, or insets, focus on different parts (e.g., faces, shoes) that can be seamlessly inserted onto the global canvas. 
We model the problem as jointly exploring the respective latent spaces such that the generated images can be combined, by inserting the parts from the specialized generators onto the global canvas, without introducing seams. We demonstrate the setup by combining a full body GAN with a dedicated high-quality face GAN to produce plausible-looking humans. We evaluate our results with quantitative metrics and user studies.
\end{abstract}
\vspace*{-.15in}

\begin{figure*}
\begin{overpic}[width=\linewidth]{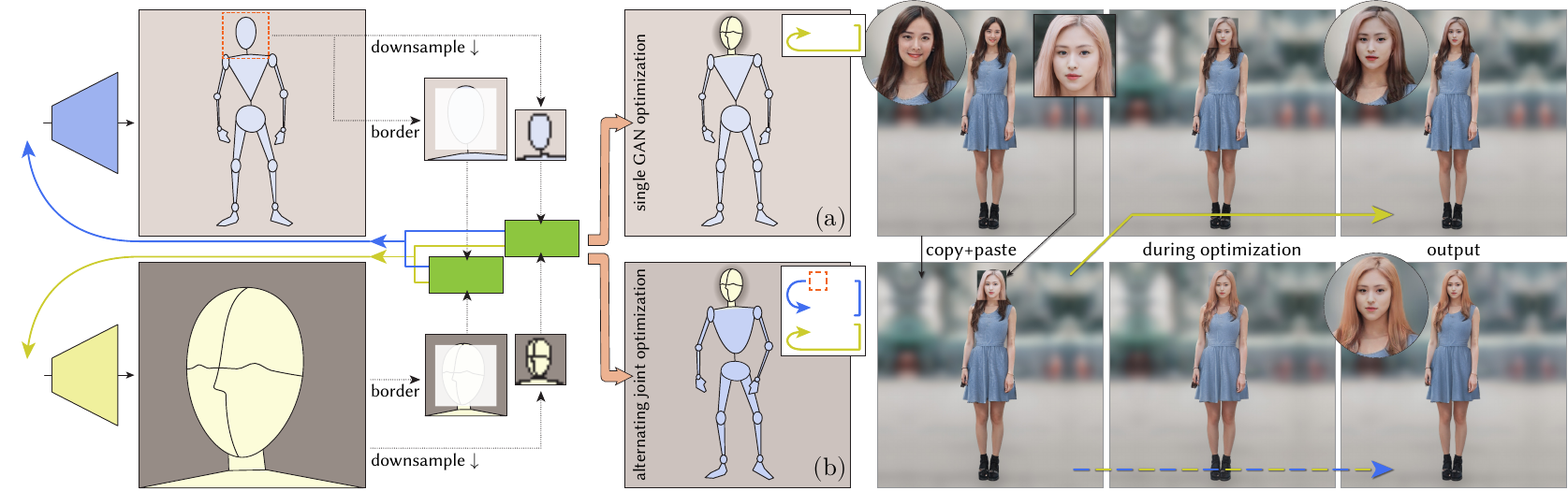}
\put(4.25,23){$\gA$}
\put(-1,23){$\wA$}
\put(9.2,28.5){$\iA$}
\put(4.25,6.8){$\gB$}
\put(-1,6.8){$\wB$}
\put(9.2,12.5){$\iB$}
\put(12.5,29){$\color{orange} \bbox$}
\put(34.9,25.6){$I^\downarrow_A$}
\put(34.9,4.4){$I^\downarrow_B$}
\put(27.7,13.5){$\scriptstyle \lB$}   
\put(32.7,15.8){$\scriptstyle \lA$}   
\put(28,24){$\border_A$}    
\put(28,7.5){$\border_B$}    
\put(51.6,28.8){$\footnotesize \wB$}
\put(53.,12.5){$\color{orange} \bbox$}
\put(51.6,11.2){$\footnotesize \wA$}
\put(51.6,9.5){$\footnotesize \wB$}

\end{overpic}
\vspace*{-0.2in}
   \caption{\textbf{\name Pipeline.} Given two latents $\wA$ and $\wB$, along with pretrained generators $\gA$ and $\gB$, that generate two images $\iA:=\gA(\wA)$ and $\iB:=\gB(\wB)$, respectively, we design a pipeline that can optimize either only $\wA$~\textbf{(a)}, or iteratively optimize both $\wA$ and $\wB$~\textbf{(b)} in order to achieve a seamless output composition of face and body. We use a set of losses $\lA$ and $\lB$ to describe the conditions we want to minimize during optimization. On the right, we show that given an input body, mere copy and pasting of a target face yields boundary artifacts. We show an application of one-way optimization \textit{(top right)} and two-way optimization \textit{(bottom right)} to create a seamlessly merged result. 
   Note that when the algorithm can optimize in both inset-face and canvas-body generator spaces, it produces more natural results at the seam boundary -- notice how the hair and skin tone blend from the head to the body region. The joint optimization is challenging as the bounding box $\bbox(\iA)$ is conditioned on the variable $\wA$.
   }
   \label{fig:pipeline}
   \vspace*{-.15in}
\end{figure*}

\vspace*{-.15in}

\section{Introduction}
\label{sec:intro}

Generative adversarial networks~(GANs) have emerged as a very successful image generation paradigm. For example, StyleGAN~\cite{Karras2020StyleGAN2} is now the method of choice for creating near photorealistic images for multiple classes (e.g., human faces, cars, landscapes). 
However, for classes that exhibit complex variations, creating very high quality results becomes harder. For example, full-body human generation still remains an open challenge, given the high variability of human pose, shape, and appearance. 

How can we generate results at both high resolution and high quality? One approach is to break the target image into tiles and train a GAN to sequentially produce them~\cite{Fruehstueck2019TileGAN}.  Such methods, however, are unsuited for cases where the coupling between the (object) parts are nonlocal and/or cannot easily be statistically modeled. An alternate approach is to aim for collecting very high resolution images and train a single GAN, at full resolution. However, this makes the data collection and training tasks very expensive, and variations in object configuration/poses cause further challenges. To the best of our knowledge, neither such a high resolution dataset, nor a corresponding high resolution GAN architecture has been published.

We propose \name towards solving the above problems. Specifically, we propose to combine a generator to provide the global context in the form of a canvas, and a set of specialized part generators that provide details for different regions of interest. The specialized results are then pasted, as insets, on to the canvas to produce a final generation. Such an approach has multiple advantages: \textbf{(I)}~the canvas GAN can be trained on medium quality data, where the object parts are not necessarily aligned. Although this results in the individual parts in the canvas being somewhat blurry (e.g., fuzzy/distorted faces in case of human bodies), this is sufficient to provide global coordination for later specialized parts to be inserted; \textbf{(II)}~the specialized parts can be trained on part-specific data, where consistent alignment can be more easily achieved; and \textbf{(III)}~different canvas/part GANs can be trained at different resolutions, thus lowering the data (quality) requirements. CollageGAN~\cite{Li2021Collage} has explored a similar idea in a conditional setting. Given a semantic map which provides useful shape and alignment hints, they create a collage using an ensemble of outputs from class-specific GANs~\cite{Li2021Collage}. In contrast, our work focuses on the unconditional setting, which is more challenging since our multiple generators need to collaborate with one another to generate a coherent shape and appearance together without access to a semantic map for hints.

The remaining problem is how to coordinate the canvas and the part GANs, such that adding the insets to the canvas does not reveal seam artifacts at the inset boundaries. This aspect is particularly challenging when boundary conditions are nontrivial and the inset boundaries themselves are unknown. For example, a face, when added to the body, should have consistent skin tone, clothing boundaries, and hair flow. We solve the problem by jointly seeking latent codes in (pretrained) canvas and part GANs such that the final image, formed by inserting the part insets on the canvas, does not exhibit any seams. 
In this paper, we investigate this problem in the context of human body generation, where the human faces are created by a face-specific GAN. 

We evaluate \name on a custom dataset, compare with alternative approaches, and evaluate the quality of the results with quantitative metrics and user studies. 
\cref{fig:teaser} shows human body generation applications highlighting both seamless results, across face insets, as well as having diversity of solutions across face insertion boundaries. 

\textbf{Contributions. }\textbf{ (1) }We propose a multi-GAN optimization framework that jointly optimizes the latent codes of two or more collaborative generators such that the overall composed result is coherent and free of boundary artifacts when the generated parts are inserted as insets into the generated canvas. \textbf{(2)} We demonstrate our framework on the highly challenging full-body human generation task and propose the first viable pipeline to generate plausible-looking humans unconditionally at 1024$\times$1024px resolution. 

\begin{figure*}
  \centering
  \includegraphics[width=0.99\linewidth]{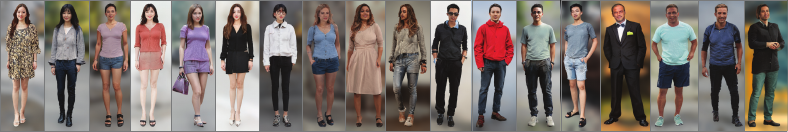}
   \vspace*{-.07in}
  \caption{\textbf{Unconditional generation results.} Examples are created using our adaptive truncation approach (described in the supplementary material) and are cropped horizontally. At first glance, the results look realistic, but face regions show noticeable artifacts (please zoom). 
  }
  \label{fig:unconditional_synthesis}
  \vspace*{-.2in}
\end{figure*}

\section{Related Work}
\label{sec:related_work}

\textbf{Unconditional Image Generation }
via Generative Adversarial Networks (GANs)~\cite{Goodfellow2014GAN} has shown a lot of promise in recent years. In this context, the StyleGAN architecture was developed over a sequence of papers~\cite{Karras2019StyleGAN, Karras2020StyleGAN2,Karras2020StyleGAN2ADA, Karras2021StyleGAN3} and is widely considered the state of the art for synthesizing individual object classes. For class-conditional image generation on the ImageNet dataset, BigGAN~\cite{Brock2018BigGAN} is often the architecture of choice. In our work, we are building on StyleGAN2-ADA, since this architecture yields better FID~\cite{Heusel2017GANs} and Precision\&Recall~\cite{Kynkaanniemi2019PnR} scores on our domain when compared to StyleGAN3. 
In addition, generating complete human body images using StyleGAN2 is a baseline we would like to improve upon in our work.

\textbf{Image Outpainting} 
refers to image completion problems where the missing pixels are not surrounded by available pixels. Recent papers build on the ideas to use generative adversarial networks~\cite{Teterwak2019ICCV} and the explicit modeling of structure~\cite{Li2021WACV,Zhao2021HumanCompletion}. Though these two papers specialize in human bodies, we find that the GAN architecture \comodgan~\cite{Zhao2021CoModGAN} has even more impressive results for image outpainting (see the comparison in  Section~\ref{sec:evaluation}). 

\textbf{Conditional Generation of Full-Body Humans} 
has two possible advantages. First, the conditional generation enables more control. Second, conditional generation can help in controlling the variability and improve the visual quality.
In the context of humans, a natural idea is to condition the generation on the human pose~\cite{Ma2017Pose,Siarohin2019PAMI,Kurupathi2020GenerationOH,Men2020CVPR,Knoche2020Reposing,Sanyal2021ICCV,AlBahar2021Pose} or segmentation information~\cite{Jiang2021BPA}.

As many conditional architectures are not able to handle the same high resolution (1024$\times$1024px) of unconditional StyleGAN, an alternative to developing new architectures is conditional embedding into an unconditional generator's latent space. Two approaches used in this context are StyleGAN embedding using optimization~\cite{abdal2019image2stylegan, abdal2020image2stylegan++} or StyleGAN embedding using an encoder architecture~\cite{richardson2021encoding, tov2021designing, alaluf2021restyle}. Our work also makes use of embedding algorithms.


\setlength{\abovedisplayskip}{5pt} \setlength{\abovedisplayshortskip}{5pt}
\setlength{\belowdisplayskip}{6pt} \setlength{\belowdisplayshortskip}{6pt}

\section{Methodology}
\label{sec:unconditional}
We propose a method for the unconditional generation of full-body human images using one or more independent pretrained unconditional generator networks. Depending on the desired application and output configuration, we describe different ways to coordinate the multiple generators. 

\subsection{Full-Body GAN}
The naive approach to generate a full-body human image is to use a single generator trained on tens of thousands of example humans (see Section~\ref{sec:dataset} about the dataset). We adopt the state-of-the-art StyleGAN2 architecture proposed by Karras et al.~\cite{Karras2020StyleGAN2ADA}. Most previous full-body generation or editing work \cite{lewis2021tryongan, AlBahar2021Pose, Zhou2021Human,Li2021Collage} generate images at 256$\times$256px or 512$\times$512px resolution. We made the first attempt to unconditionally generate full-body humans at 1024$\times$1024px resolution. Due to the complex nature of our target domain, the results generated by a single GAN sometimes exhibit artifacts such as weirdly-shaped body parts and non-photorealistic appearance. These artifacts are most visible in faces and extremities, as shown in \cref{fig:teaser}(a). Due to the vast diversity of human poses and appearances and the associated alignment difficulty, hands and feet appear in many possible locations in the training images, making them harder for a single generator to learn. Faces are especially hard since we humans are ultra-sensitive to artifacts in these areas. They therefore deserve dedicated networks and special treatment. \cref{fig:unconditional_synthesis} shows a variety of unconditional generation results. Our results exhibit correct human body proportions, consistent skin tones across face and body, interesting garment variations and plausible-looking accessories (e.g. handbags and sunglasses) whereas artifacts can be present when viewed in detail.

\subsection{Multi-GAN Optimization}
To improve the problematic regions generated by the full-body GAN, we use other generators trained on images of specific body regions to generate pixels to be pasted, as insets, into the full-body GAN result. The base full-body GAN and the dedicated body part GANs can be trained using the same or different datasets. In either case, the additional network capacity contained in the multiple GANs can better model the complex appearance and variability of the human bodies. 

\begin{figure}[ht]
  \centering
   \includegraphics[width=0.99\linewidth]{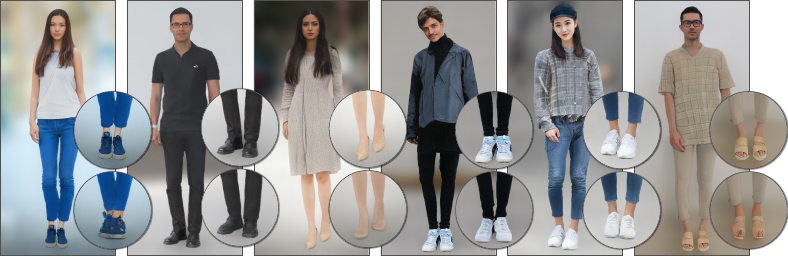}
   \vspace*{-.07in}
   \caption{\textbf{Two Insets}. These results are improved using a dedicated shoe generator trained on shoe crops from our full-body humans, and also using our face generator. All three generators (full-body canvas and two insets) are jointly optimized to produce a seamless coherent output. The circular closeups show the shoes before \textit{(top)} and after \textit{(bottom)} improvement (please zoom). 
   }
   \label{fig:two_insets}
   \vspace*{-.2in}
\end{figure}

As a proof of concept, we show that a face GAN trained with the face regions cropped from our full-body training images can be used to improve the appearance of the body GAN results. Alternatively, we can also leverage a face generator trained on other datasets such as FFHQ~\cite{Karras2020StyleGAN2} for face enhancement as well. 
Similarly, specialized hands or feet generators can also be used in our framework to improve other regions of the body. We show that we can also use multiple part generators together in a multi-optimization process, as depicted in \cref{fig:two_insets}.

The main challenge is how to coordinate multiple unconditional GANs to produce pixels that are coherent with one another. In our application, we have a $\gA$ that generates the full-body human where $\iA:=\gA(\wA)$ and another $\gB$ that generates a sub-region or inset within the human body where $\iB:=\gB(\wB)$. In order to coordinate the specialized part GAN with the global/canvas GAN, we need a bounding box detector to identify the region of $\iA$ that corresponds to the region our part GAN generates. We crop $\iA$ with the detected bounding box and denote the cropped pixels as $\bbox(\iA)$.  
The problem of inserting a separately-generated part $\iB$ into the canvas $\iA$ is equivalent to finding a latent code pair $(\wA, \wB)$ such that the respective images $\iA$ and $\iB$ can be combined without noticeable seams in the boundary regions of $\bbox(\iA)$ and $\iB$. To generate the final result, we directly replace the original pixels inside the bounding box $\bbox(\iA)$ with the generated pixels from $\iB$, 

\vspace*{-.1in}
\begin{equation}
\min_{\wA, \wB} \int_{\Omega} \mathcal{L}( \gA(\wA), \gB(\wB )) 
\label{eqn:optimization}
\end{equation}
where, $\Omega := \bbox(\gA(\wA))$ and, with slight abuse of notation, $\mathcal{L}$ captures the loss both along the boundary of $\Omega$ measuring seam quality and inside the region $\Omega$ measuring similarity of $\iA$ and $\iB$ inside the respective faces. The full optimization is complex as the region of interest $\Omega$ depends on $\wA$. 

Our multi-GAN optimization framework can support various human generation and editing applications. Depending on the application scenario, we optimize either $\wA$ or $\wB$ or jointly optimize both for the best results. 

\textbf{Optimization Objectives. }
When optimizing the latent codes $\wA$, $\wB$ or both, we consider multiple objectives: \textbf{(I)}~the face regions generated by the face GAN and body GAN should have similar appearance at a coarse scale so that when the pixels generated by the face GAN are pasted onto the body GAN canvas, attributes match (e.g., the skin tone of the face matches that of the neck); \textbf{(II)}~the boundary pixels around the face crops match up so that a simple copy-and-paste operation does not result in visible seams; and \textbf{(III)}~the final composed result looks realistic. 
\begin{figure}[ht]
  \centering
   \includegraphics[width=0.99\linewidth]{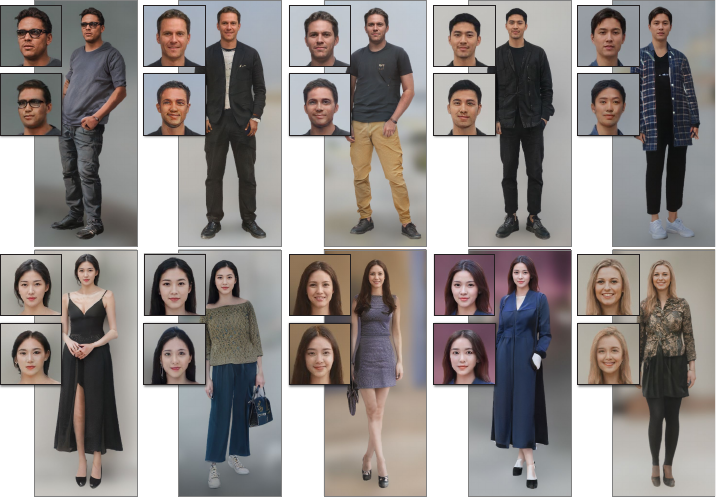}
   \vspace*{-.07in}
   \caption{\textbf{Face Refinement}. Given generated humans, we use a dedicated face model trained on the same dataset to improve the quality of the face region. We jointly optimize both the face and the human latent codes so that the two generators coordinate with each other to produce coherent results. The two inset face crops show the initial face generated by the body GAN \textit{(bottom)} and the final face improved by dedicated face GAN \textit{(top)}.
   }
   \label{fig:face_enhancement}
   \vspace*{-.2in}
\end{figure}
To match the face appearance, we downsample the face regions and calculate a combination of $L1$ and perceptual loss~\cite{zhang2018perceptual} $\percep$: 
\begin{equation}
\lA := \weight_1\Lone(\dA, \dB) + \weight_2\percep(\dA, \dB), 
\end{equation}
where $\dA = \downsample_{64}(\bbox(\iA))$ and $\dB = \downsample_{64}(\iB)$ and $\downsample_{64}$ refers to downsampling the image to 64$\times$64px resolution. 

For the boundary matching, we also apply a $L1$ and perceptual loss to the border pixels at full resolution:
\begin{equation}
\begin{split}
\lB := \weight_3\Lone(\border_8(\bbox(\iA)), \border_8(\iB)) + \\ \weight_4\percep(\border_8(\bbox(\iA)), \border_8(\iB))
\end{split}
\end{equation}
where $\border_x(I)$ is the border region of $I$ of width $x$ pixels. 

To maintain realism during the optimization, we also add two regularization terms:
\begin{equation}
\vspace*{-.1in}
\lR := \weight_{r1}\|\w^* - \wAvg\| + \weight_{r2} \sum_{i}\|\delta_i\|
\end{equation}
The first term prevents the optimized latent code from deviating too far from the average latent.
We compute $\wAvg$ by randomly sampling a large number of latents in $Z$ space, mapping them to $W$ space, and computing the average.
The second term is to regularize the latent code in $\w^{+}$ latent space. During StyleGAN2 inference, the same $512$ dimensional latent code $\w$ is fed into each of the $n$ generator layers ($n$ is dependent on the output resolution). Many GAN inversion methods optimize in this $n \times 512$ dimensional $\w^{+}$ latent space~\cite{Xia2021InversionSurvey} instead of the $512$ dimensional $\w$ latent space. We follow recent work to decompose the $\w^{+}$ latent into a single base $\w^*$ latent and $n$ offset latents $\delta_i$. The latent used for layer $i$ is $\w + \delta_i$. We use the $L_2$ norm as regularizer to ensure that the $\delta_i$s remain small.
Based on our visual analysis of the results, we use larger weights for the body generator than the face generator for this regularizer.

We mix and match the various losses depending on the specific application at hand. 

\begin{figure}[t]
  \centering
   \includegraphics[width=0.99\linewidth]{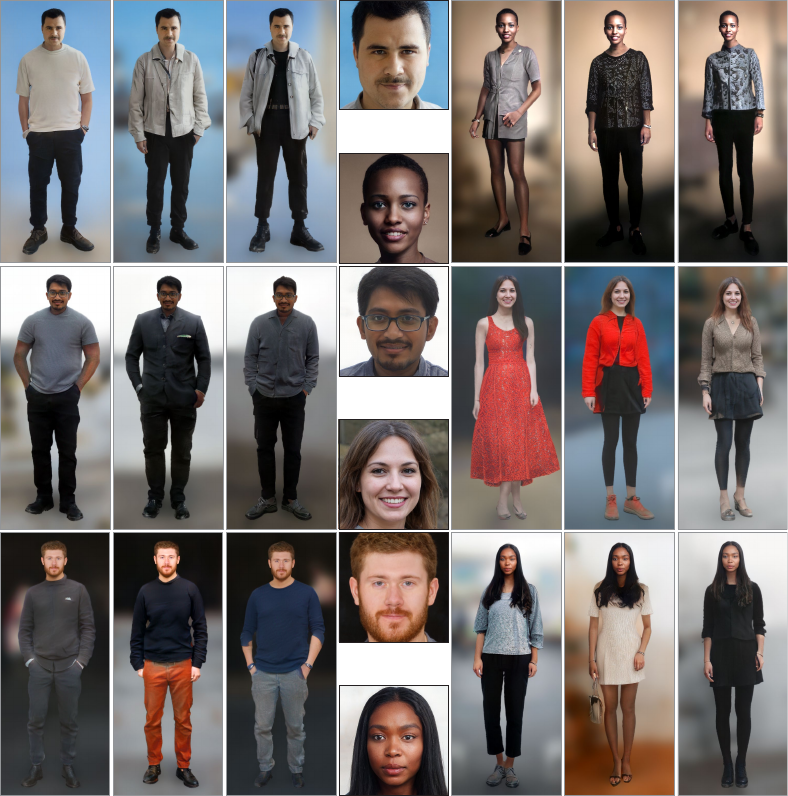}
   \vspace*{-.07in}
   \caption{\textbf{Multimodal Body Generation for an Existing Face.} For each face generated by the pretrained FFHQ model \textit{(middle column)}, we use joint optimization to generate three different bodies while maintaining the facial identities from the input faces.
   }
   \label{fig:body_generation}
   \vspace*{-.15in}
\end{figure}

\textbf{Face Refinement versus Face Swap. } 
Given a randomly generated human body $\gA(\wA)$, we can keep $\wA$ fixed and optimize for $\wB$ such that $\gB(\wB)$ looks similar to $\bbox(\gA(\wA))$ at a coarse scale and matches the boundary pixels at a fine scale (\cref{fig:pipeline} top right). We have:
\begin{equation}
\min_{\wB} (\lA + \lB)
\end{equation}
While this almost produces satisfactory results, boundary discontinuities show up at times. For further improvement, we can optimize both $\wA$ and $\wB$ so that both generators coordinate with each other to generate a coherent image free of blending artifacts (\cref{fig:pipeline} bottom right). To keep the body appearance unchanged during the optimization of $\wA$, we introduce an additional loss term:
\begin{equation}
\begin{split}
\lRB := \weight_5\Lone(\region^O(\iA), \region^O(\iRef)) + \\ \weight_6\percep(\region^O(\iA), \region^O(\iRef)) 
\end{split}
\end{equation}
where $\iRef$ is the input reference body generated by $\gA$ that should remain unchanged during the optimization, $\region^O$ defines the body region outside of the face bounding box. We also use the mean latent regularization term $\lR$ to prevent generating artifacts. 
The final objective function becomes:
\begin{equation}
\min_{\wA,\wB} (\lA + \lB + \lR + \lRB)
\label{eq1}
\end{equation}
\cref{fig:teaser}(b) and \cref{fig:face_enhancement} show face refinement results using a dedicated face model trained on faces cropped from the same data used for training the body generator. Our refinement results when using the pretrained FFHQ face model exhibit similar visual quality (see supplementary material).

\textbf{Body Generation for an Existing Face. } 
Given a real face or a randomly-generated face $\gB(\wB)$, we can keep $\wB$ fixed and optimize for $\wA$ such that $\gA(\wA)$ produces a body that looks compatible with the input face in terms of pose, skin tone, gender, hair style, etc. In practice, we find that to best maintain boundary continuity, especially when generating bodies to match faces of complex hair styles, it is often to discourage large changes in $\wB$, such that the face identity is mostly preserved but the boundary and background pixels can be slightly adjusted to make the optimization of $\wA$ easier. To preserve the face identity during the optimization, we use an additional face reconstruction loss:
\begin{equation}
\begin{split}
\lRF := \weight_7\Lone(\region^I(\iB), \region^I(\iRef)) + \\\weight_8\percep(\region^I(\iB), \region^I(\iRef))
\end{split}
\end{equation}
where $\region^I$ defines the interior region of the face crop and $\iRef$ denotes the referenced input face. For more precise control, face segmentation can be used instead of bounding boxes.
Our objective function becomes: 
\begin{equation}
\min_{\wA,\wB} (\lA + \lB + \lR + \lRF)
\end{equation}
With different initialization for $\wA$, we can generate multiple results per face as shown in \cref{fig:body_generation}. Note that our model can generate diverse body appearances compatible with the input face. The generated body skin tone generally match the input face skin tone (e.g., the women of African descent in the top and bottom rows of \cref{fig:body_generation}). \cref{fig:teaser}(c) shows another example.

\begin{figure}[t]
  \centering
   \includegraphics[width=0.99\linewidth]{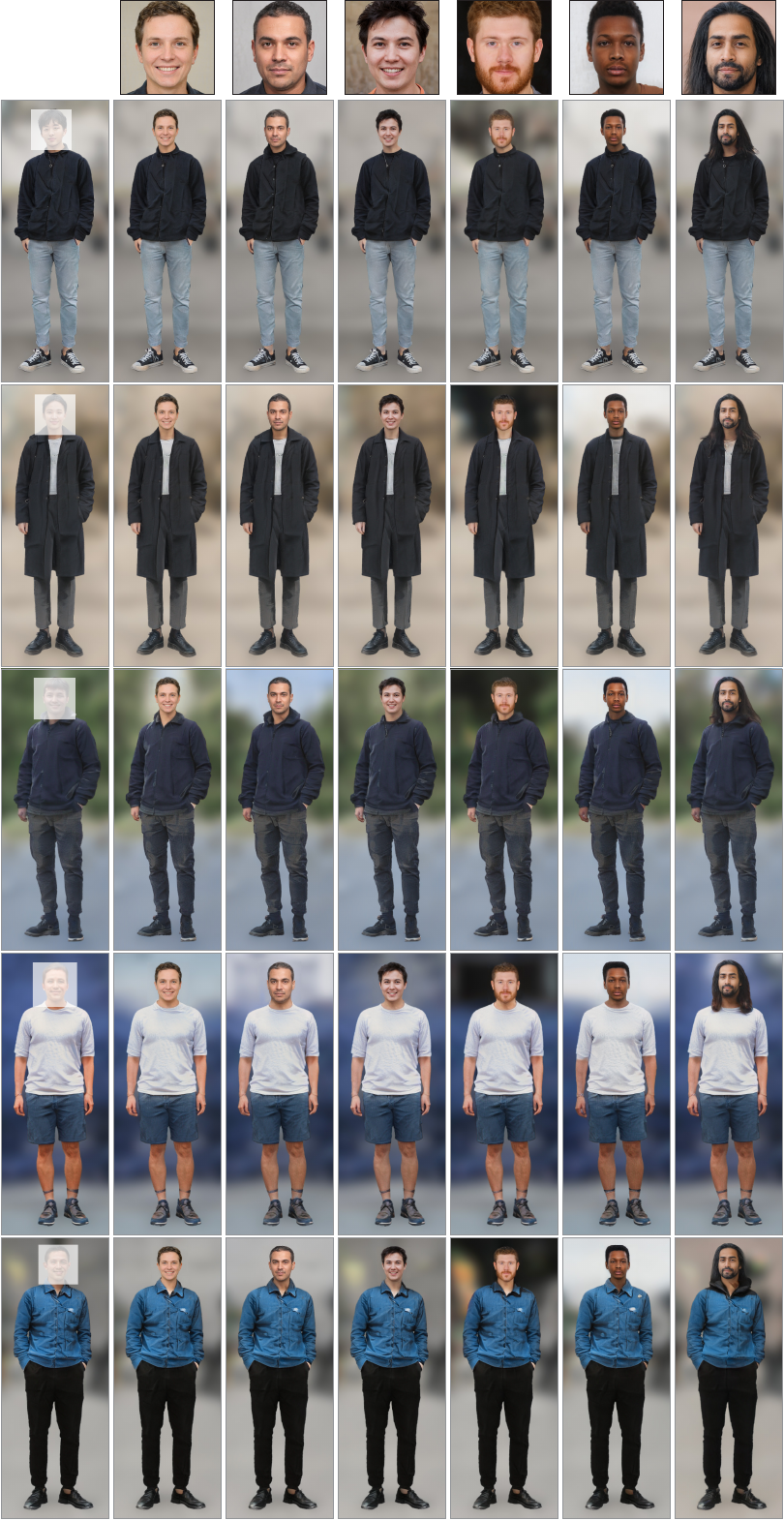}
   \vspace*{-.07in}
   \caption{\textbf{Face Body Montage.} Given target faces \textit{(top row)} generated by the pretrained FFHQ model and bodies \textit{(leftmost column)} generated by our full-body human generator, we apply \textit{joint} latent optimization to find compatible face and human latent codes that can be combined to produce coherent full-body humans. Notice how face and skin colors get synchronized, and zoom in to observe the (lack of) seams around face insets. 
   }
   \label{fig:face_clothing_matrix}
   \vspace*{-.23in}
\end{figure}

\textbf{Face Body Montage. } 
We can combine any real or generated face with any generated body to produce a photo montage. With a real face, we need to first encode it into the latent space of $\gB$ as $\wB$ using an off-the-shell encoder~\cite{tov2021designing}. Similarly, a real body could be encoded into the latent space of $\gA$, but due to the high variability of human bodies it is difficult to achieve a low reconstruction error.
All montage results are created from synthetic bodies generated by $\gB$. We use the following objective function:
\begin{equation}
\min_{\wA,\wB} (\lA + \lB + \lR + \lRF + \lRB) 
\end{equation}
\cref{fig:face_clothing_matrix} shows the result of combining faces \textit{(top row)} generated by the pretrained FFHQ model with bodies \textit{(leftmost column)} generated by our full-body generator $\gA$. With minor adjustments of both the face and body latent codes, we achieve composition results that are coherent and identity-preserving. While we do not have any explicit loss encouraging skin tone coherence, given faces with different skin tones, our joint optimization slightly adjusts the skin tone of the body's neck and hand pixels to minimize appearance incoherence and boundary discrepancy in the final results. \cref{fig:teaser}(d) shows two more examples. Our joint optimization is able to slightly adjust the shoulder region of the lady to extend her hair to naturally rest on her right shoulder. The rightmost column in \cref{fig:pipeline} shows the improvement the joint optimization makes to the final result quality \textit{(bottom)} compared to only optimizing $\wB$ given an input body \textit{(top)}. 

\begin{figure}[t]
 \centering
  \includegraphics[width=0.95\linewidth]{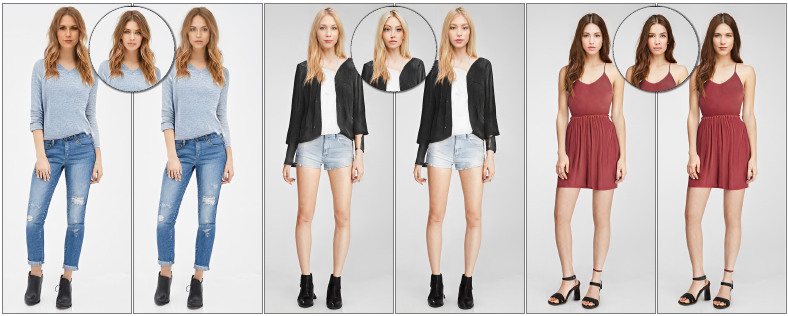}
   \vspace*{-.07in}
  \caption{\textbf{Multimodal Face Improvement.} To improve humans generated by a full-body model trained on DeepFashion, we use the pretrained FFHQ model to synthesize a variety of seamlessly merged result faces that all look compatible with the input body. }
  \label{fig:face_deepfashion}
  \vspace*{-.15in}
\end{figure}

\textbf{Optimization Details. } 
While the difference is subtle, we observe a slightly better visual performance when using $\Lone$ over $\Ltwo$ losses. We apply many of our losses to downsampled versions of the images $\downsample_{64}(\bbox(\iA))$ and $\downsample_{64}(\iB)$ to allow for more flexibility during optimization and to reduce the risk of overfitting to artifacts from the source image (e.g., the body GAN's face region, which lacks realistic high-frequency details) in a strategy similar to PULSE~\cite{Menon2020Pulse}.

One challenge in the joint optimization of $\wA$ and $\wB$ is that the boundary condition $\Omega$ depends on the variable $\wA$. We address this by alternately optimizing for $\wA$ and $\wB$, and reevaluating the boundary after each update of $\wA$. We stop the process when the updates converge. 

\textbf{Optimization Initialization.} The default choice of initialization for either $\wA$ or $\wB$ is their corresponding average latent vector $\wAvg$. This typically leads to reasonable results quickly. However, it is desirable to generate a variety of results for applications like finding matching bodies $\iA$ for an input face $\iB$. In this case, we start from truncated latent codes $\wTrunc = \wRand * (1 - \alpha) + \wAvg * \alpha$. Due to the introduced randomness and the interpolation with the average latent code, we can generate diverse yet realistic results (see \cref{fig:body_generation}). In \cref{fig:face_deepfashion}, given humans generated by our full-body model trained on DeepFashion, we use the pretrained FFHQ face model to swap in multiple better looking faces. Different initialization of $\wB$ yields different results. In the cases where either the face region or the body region should remain fixed during the joint optimization of both latent codes, we initialize the optimization with the latent code initially used to generate the synthetic reference image or the latent code encoded from a real image.


\section{Dataset and Implementation}
\label{sec:dataset}

\begin{figure*}
  \centering
  \includegraphics[width=0.99\linewidth]{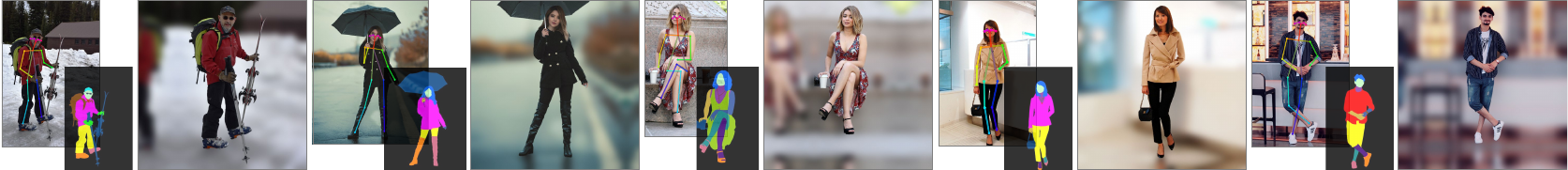}
   \vspace*{-.07in}
  \caption{\textbf{Full-body Human Dataset.} 
  We create a dataset from photographs of humans in the wild. The images are automatically preprocessed, aligned, and cropped to 1024$\times$1024px resolution using the ground-truth segmentation masks and detected pose skeletons.} 
  \label{fig:dataset}
  \vspace*{-.15in}
\end{figure*}

 We curate a proprietary dataset of 83,972 high-quality full-body human photographs at 1024$\times$1024px resolution. These images stem from a dataset of 100,718 diverse photographs in the wild purchased from a third-party data vendor. The dataset includes hand-labeled ground-truth segmentation masks. We apply a human pose-detection network~\cite{Cao2021OpenPose} on the original images and filter out those that contain extreme poses causing pose detection results to have low confidence. \cref{fig:dataset} shows some sample training images.

Feature alignment plays an important role in high-quality image generation, as can be seen in the qualitative difference of models trained on FFHQ data vs.\ other face datasets. Therefore, we carefully align the humans using their pose skeletons. We define an upper body axis based on the position of the neck and hip joints. We position humans so that the upper body axis is aligned in the center of the image. As the variance in perspective and pose is very large, choosing the appropriate scale for each person within the context of their image frame is challenging. We scale the humans based on their upper-body length and then evaluate the extent of the face region as defined by the segmentation mask. If the face length is smaller (larger) than a given minimum (maximum) value, we rescale so that the face length is equal to the minimum (maximum). 

Lastly, we enlarge the backgrounds using reflection padding and heavily blur them using a Gaussian kernel of size 27 to focus the generator capacity on modeling only the foreground humans. The huge variation in background appearance in these in-the-wild photos poses extreme challenges for the GAN, especially with limited data quantity. 

We also considered completely removing the background, but did not do it for two reasons: \textbf{(1)} Human-labeled segmentation masks are still imperfect around boundaries and \textbf{(2)} We observe that current GAN architectures do not handle large areas of uniform color well.

We also show our method on DeepFashion~\cite{Liu2016DeepFashion}, which consists of 66,607 fashion photographs, including garment pieces and garments on humans.
Using the same alignment strategy as above, we extract 10,145 full-body images at 1024$\times$768 resolution. Since the backgrounds are already uniform, we do not blur them.

\textbf{Training Details} 
We trained our main human body generator network at 1024$\times$1024px resolution using the StyleGAN2-ADA architecture using all augmentation schemes proposed in the paper~\cite{Karras2020StyleGAN2ADA} for 28 days and 18 hours on 4 Titan V GPUs, using a batch size of 4, processing a total of 42M images.
After experimenting with different $R1~\gamma$ values between 0.1 and 20, we chose a value of 13. Similarly, we trained our DeepFashion human generator network at 1024$\times$768px resolution for 9 days on 4 v100 GPUs, using a batch size of 8, processing a total of 18M images. We use a pretrained FaceNet~\cite{Schroff2015FaceNet} to detect and align the bounding boxes of the face regions in our generated bodies and faces. The running time of our optimization algorithm for jointly optimizing two generator latents at 1024$\times$1024px output resolution is about 75 seconds on a Titan RTX GPU. If $\gB$ has a smaller resolution of 256$\times$256px, the optimization time decreases to around 60 seconds.

\section{Evaluation and Discussion}
\label{sec:evaluation}

\textbf{Quantitative Evaluation.}
We follow the standard practice to calculate FID (Fréchet Inception Distance) to measure how closely our generated full-body results follow the training distribution. Many previous papers including \comodgan point out that FID statistics are noisy and do not correlate well with human perception about visual quality. We also observe that FID is more sensitive to result diversity than quality and increases significantly as we truncate the generated results, which reduces variation but is crucial for generating natural looking images with fewer artifacts. While the FID for untruncated results is 13.96, it rises to 26.67 for t=0.7 and 71.90 for t=0.4 (more truncation).
We compare FID values of several alternative approaches for our face refinement application. We use two different truncation settings, t=0.7 and t=0.4 and evaluate on both the full-body images and image crops that include the refined face and the boundary pixels after copy\&pasting.

\vspace*{-.15in}
\begin{center}
\small
\begin{tabular}{*5c}
\toprule
 &  \multicolumn{2}{c}{t=0.7} & \multicolumn{2}{c}{t=0.4}\\
{FID Score (lower is better)}                          & body   & face    & body   & face\\
\midrule
Unconditional Generation    & 26.67  & \textbf{27.14}   & 71.90 & 66.61\\
\name                       & \textbf{25.33}  & 31.61   & \textbf{69.58} & \textbf{61.57}\\
\bottomrule
\end{tabular}
\end{center}

The differences in FID are small. This indicates that the face refinement using joint optimization does not modify the distribution learned by the unconditional generator and therefore does not decrease the result diversity. However, large differences in perceptual quality are still possible despite similar FID values as demonstrated in our user study.

\begin{figure}[t]
  \centering
   \includegraphics[width=0.99\linewidth]{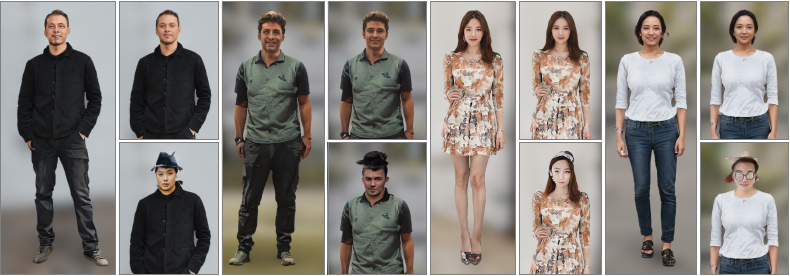}
   \vspace*{-.07in}
   \caption{\textbf{Face Refinement Comparison with \comodgan~\cite{Zhao2021CoModGAN}.} Given generated humans \textit{(left)}, \name improves the face quality \textit{(top right)}, producing more convincing results than \comodgan \textit{(bottom right)}. \comodgan results are generated by defining rectangular holes around the face regions.
   }
   \label{fig:comod_face}
   \vspace*{-.22in}
\end{figure}

\begin{figure}[t]
  \centering
   \includegraphics[width=0.99\linewidth]{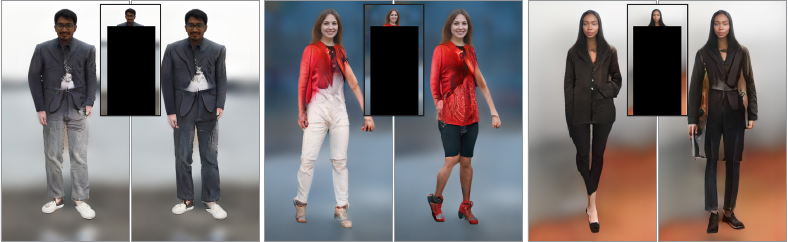}
   \vspace*{-.07in}
   \caption{\textbf{Body Generation with \comodgan~\cite{Zhao2021CoModGAN}.} We show results generated by \comodgan trained to fill a rectangular hole covering the body in a given image. Inputs with holes are shown in insets. We generate several results per input and show the best looking two here. Please refer to \cref{fig:body_generation} for our results on the same input faces. We observe that \comodgan creates seamless content, but worse visual quality compared to ours.
   }
   \label{fig:comod_body}
   \vspace*{-.22in}
\end{figure}

\textbf{Baseline Comparison. } 
To the best of our knowledge, no other prior work generates full-body humans unconditionally or inpaints/outpaints humans at 1024$\times$1024px resolution without requiring conditioning other than reference pixels of the known regions. Previous works have attempted to generate plausible human bodies, but they require segmentation masks as input~\cite{Li2021Collage, Li2021WACV}. The best state-of-the-art method that can be repurposed for our body generation and face refinement applications is \comodgan~\cite{Zhao2021CoModGAN}. \cref{fig:comod_face} shows that our \name \textit{(top right)} outperforms \comodgan (bottom right) in replacing the initial face generated by our body generator \textit{(left)}. We trained \comodgan with square (with small random offsets for generalization) holes around faces using the official implementation, training data and default parameters for two weeks on four V100 GPUs. Similarly, we train \comodgan with rectangular holes around the bodies to compare with our \name for the body generation task. In \cref{fig:comod_body}, we show the two best results of \comodgan obtained by using several random initializations per input face. Compared to our results in \cref{fig:body_generation}, \comodgan produces less realistic and diverse image completions.

\textbf{User Study. }
We performed a user study to better evaluate the perceptual quality of our method. We aggregated 500 generated humans from our full-body generator and 500 random training images. We then applied either our joint optimization method or \comodgan to replace the face regions in the generated samples. We showed several sets of image pairs to volunteer participants on Amazon Mechanical Turk and asked them to pick ``in which of the two images the person looks more plausible and real". Per image pair, 5 votes were collected and aggregated towards the majority votes.
The study shows that in 12.4\% of image pairs, users prefer our unrefined results over the training images when given only 1 second to look at each image. This shows that our results get the basic human proportion and pose right, being able to confuse people about being real. In 98\% of cases, users prefer our joint optimization results over the unrefined images. In contrast, only 7\% of the \comodgan samples were picked over the unrefined images, which is consistent with our observation from \cref{fig:comod_face}. 

\textbf{Limitations.}
Our work has multiple limitations that could benefit from improvements. First, the joint optimization approach may change details such as hair style, neckline or clothing details. In many cases the changes are minor, but in some cases changes can be larger, e.g. the woman's hair in the middle row of \cref{fig:body_generation} and the man's collar in the top row of \cref{fig:face_clothing_matrix}.  Second, as shown in \cref{fig:unconditional_synthesis}, our full-body GAN has other problems not improved by \name: symmetry, e.g. noticeable in the hands and feet and on outfits (the woman with a light blue shirt), and the consistency of the fabric used for the clothing (the rightmost person). Third, the generated results have limited variations in body type and pose as discussed next in more detail. The vast majority of our generated results have a slender body type due to the training data distribution.

\textbf{Dataset Bias and Societal Impact.} 
Both DeepFashion and our proprietary dataset contain biases. DeepFashion contains a limited number of unique identities. The same models appear in multiple images. An overwhelming majority of the images are female (around 9:1) and the range of age, ethnicity and body shape does not represent the real human population. As a result, models trained on it can only generate limited range of identities (mostly young white females) as shown in \cref{fig:face_deepfashion}. We made our best attempt to look for a diverse dataset and purchased one from a data vendor in East Asia but noticed the over-representation of young Asian females in the images. Also, many images appear to portray slim street fashion models; as a result the vast majority of them contain slender body type and formal attires. Models trained on biased datasets tend to learn biased representations of the human body. Due to the over-representation of Asians, our results on other ethnicities contain more artifacts in the face region (see the rightmost four results in \cref{fig:unconditional_synthesis}). We encourage future research efforts to diversify the training dataset to better serve our diverse society. The unconditional nature of our generation process combined with the limitation of some optimization losses operating at a low resolution implies that our generated human outputs might not preserve the attributes in the input image exactly. In addition, since the age distribution in our data is almost exclusively adult humans, we are not able to faithfully produce bodies for faces of children. Similar to other human domain generative models, our approach can be exploited by malicious users to produce deep fakes. However, as we have seen in the user study, even in a short second, users identify most of our generated results as fake compared to real human images. As we further improve the result qualities, we hope, and encourage other researchers, to investigate deep fake detection algorithms.

\section{Conclusion}
\label{sec:conclusion}

We presented \name, the first viable framework to generate plausible-looking human images unconditionally at 1024$\times$1024px resolution. The main technical contribution of \name is to introduce a multi-GAN optimization framework that jointly optimizes the latent codes of two or more collaborative generators.
In future work, we propose to extend the multi-generator idea to 3D shape representations, such as 3D GANs or auto-regressive models based on transformers. We also plan to demonstrate the \name framework on other image domains and investigate coordinated latent editing in the multi-GAN setup. 

{
\small
\bibliographystyle{ieee_fullname}
\bibliography{human_synthesis_new}

\begin{thebibliography}{10}\itemsep=-1pt

\bibitem{abdal2019image2stylegan}
Rameen Abdal, Yipeng Qin, and Peter Wonka.
\newblock {Image2StyleGAN}: How to embed images into the stylegan latent space?
\newblock In {\em Proceedings of the IEEE/CVF International Conference on
  Computer Vision and Pattern Recognition (CVPR)}, pages 4432--4441, 2019.

\bibitem{abdal2020image2stylegan++}
Rameen Abdal, Yipeng Qin, and Peter Wonka.
\newblock {Image2StyleGAN++}: How to edit the embedded images?
\newblock In {\em Proceedings of the IEEE/CVF International Conference on
  Computer Vision and Pattern Recognition (CVPR)}, pages 8296--8305, 2020.

\bibitem{alaluf2021restyle}
Yuval Alaluf, Or Patashnik, and Daniel Cohen-Or.
\newblock Restyle: A residual-based stylegan encoder via iterative refinement.
\newblock In {\em Proceedings of the IEEE/CVF International Conference on
  Computer Vision and Pattern Recognition (CVPR)}, pages 6711--6720, 2021.

\bibitem{AlBahar2021Pose}
Badour AlBahar, Jingwan Lu, Jimei Yang, Zhixin Shu, Eli Shechtman, and Jia-Bin
  Huang.
\newblock Pose with {S}tyle: {D}etail-preserving pose-guided image synthesis
  with conditional stylegan.
\newblock {\em ACM Transactions on Graphics}, 2021.

\bibitem{Brock2018BigGAN}
Andrew Brock, Jeff Donahue, and Karen Simonyan.
\newblock Large scale {GAN} training for high fidelity natural image synthesis.
\newblock In {\em International Conference on Learning Representations (ICLR)},
  2019.

\bibitem{Cao2021OpenPose}
Zhe Cao, Gines Hidalgo, Tomas Simon, Shih-En Wei, and Yaser Sheikh.
\newblock {OpenPose}: Realtime multi-person {2D} pose estimation using part
  affinity fields.
\newblock {\em IEEE Transactions on Pattern Analysis \& Machine Intelligence},
  43(01):172--186, 2021.

\bibitem{Fruehstueck2019TileGAN}
Anna Fr\"{u}hst\"{u}ck, Ibraheem Alhashim, and Peter Wonka.
\newblock {TileGAN}: Synthesis of large-scale non-homogeneous textures.
\newblock {\em ACM Transactions on Graphics (Proceedings of ACM SIGGRAPH)},
  38(4):58:1--58:11, 2019.

\bibitem{Goodfellow2014GAN}
Ian Goodfellow, Jean Pouget-Abadie, Mehdi Mirza, Bing Xu, David Warde-Farley,
  Sherjil Ozair, Aaron Courville, and Yoshua Bengio.
\newblock Generative adversarial nets.
\newblock {\em Advances in neural information processing systems}, 27, 2014.

\bibitem{Heusel2017GANs}
Martin Heusel, Hubert Ramsauer, Thomas Unterthiner, Bernhard Nessler, and Sepp
  Hochreiter.
\newblock {GANs} trained by a two time-scale update rule converge to a local
  nash equilibrium.
\newblock {\em Advances in neural information processing systems}, 30, 2017.

\bibitem{Jiang2021BPA}
Jinfeng Jiang, Guiqing Li, Shihao Wu, Huiqian Zhang, and Yongwei Nie.
\newblock {BPA}-{GAN}: Human motion transfer using body-part-aware generative
  adversarial networks.
\newblock {\em Graphical Models}, 115:101107, 2021.

\bibitem{Karras2020StyleGAN2ADA}
Tero Karras, Miika Aittala, Janne Hellsten, Samuli Laine, Jaakko Lehtinen, and
  Timo Aila.
\newblock Training generative adversarial networks with limited data.
\newblock In {\em Proceedings of the IEEE Conference on Neural Information
  Processing Systems (NeurIPS)}, 2020.

\bibitem{Karras2021StyleGAN3}
Tero Karras, Miika Aittala, Samuli Laine, Erik H\"ark\"onen, Janne Hellsten,
  Jaakko Lehtinen, and Timo Aila.
\newblock Alias-free generative adversarial networks.
\newblock In {\em Proc. NeurIPS}, 2021.

\bibitem{Karras2019StyleGAN}
Tero Karras, Samuli Laine, and Timo Aila.
\newblock A style-based generator architecture for generative adversarial
  networks.
\newblock In {\em Proceedings of the IEEE/CVF International Conference on
  Computer Vision and Pattern Recognition (CVPR)}, pages 4401--4410, 2019.

\bibitem{Karras2020StyleGAN2}
Tero Karras, Samuli Laine, Miika Aittala, Janne Hellsten, Jaakko Lehtinen, and
  Timo Aila.
\newblock Analyzing and improving the image quality of {StyleGAN}.
\newblock In {\em Proceedings of the IEEE/CVF International Conference on
  Computer Vision and Pattern Recognition (CVPR)}, pages 8107--8116, 2020.

\bibitem{Knoche2020Reposing}
Markus Knoche, Istvan Sarandi, and Bastian Leibe.
\newblock Reposing humans by warping {3D} features.
\newblock In {\em Proceedings of the IEEE/CVF Conference on Computer Vision and
  Pattern Recognition (CVPR) Workshops}, June 2020.

\bibitem{Kurupathi2020GenerationOH}
Sheela~Raju Kurupathi, Pramod Murthy, and Didier Stricker.
\newblock Generation of human images with clothing using advanced conditional
  generative adversarial networks.
\newblock In {\em DeLTA}, 2020.

\bibitem{Kynkaanniemi2019PnR}
Tuomas Kynk{\"a}{\"a}nniemi, Tero Karras, Samuli Laine, Jaakko Lehtinen, and
  Timo Aila.
\newblock Improved precision and recall metric for assessing generative models.
\newblock {\em Advances in Neural Information Processing Systems}, 32, 2019.

\bibitem{lewis2021tryongan}
Kathleen~M Lewis, Srivatsan Varadharajan, and Ira Kemelmacher-Shlizerman.
\newblock {TryOnGAN}: Body-aware try-on via layered interpolation.
\newblock {\em ACM Transactions on Graphics (Proceedings of ACM SIGGRAPH)},
  40(4), 2021.

\bibitem{Li2021WACV}
Yijun Li, Lu Jiang, and Ming-Hsuan Yang.
\newblock Controllable and progressive image extrapolation.
\newblock In {\em Proceedings of the IEEE/CVF Winter Conference on Applications
  of Computer Vision (WACV)}, pages 2140--2149, January 2021.

\bibitem{Li2021Collage}
Yuheng Li, Yijun Li, Jingwan Lu, Eli Shechtman, Yong~Jae Lee, and Krishna~Kumar
  Singh.
\newblock Collaging class-specific {GANs} for semantic image synthesis.
\newblock In {\em Proceedings of the IEEE/CVF International Conference on
  Computer Vision (ICCV)}, October 2021.

\bibitem{Liu2016DeepFashion}
Ziwei Liu, Ping Luo, Shi Qiu, Xiaogang Wang, and Xiaoou Tang.
\newblock {DeepFashion}: Powering robust clothes recognition and retrieval with
  rich annotations.
\newblock In {\em Proceedings of the IEEE/CVF International Conference on
  Computer Vision and Pattern Recognition (CVPR)}, June 2016.

\bibitem{Ma2017Pose}
Liqian Ma, Xu Jia, Qianru Sun, Bernt Schiele, Tinne Tuytelaars, and Luc
  Van~Gool.
\newblock Pose guided person image generation.
\newblock In {\em Advances in Neural Information Processing Systems}, pages
  405--415, 2017.

\bibitem{Men2020CVPR}
Yifang Men, Yiming Mao, Yuning Jiang, Wei-Ying Ma, and Zhouhui Lian.
\newblock Controllable person image synthesis with attribute-decomposed {GAN}.
\newblock In {\em Proceedings of the IEEE/CVF International Conference on
  Computer Vision and Pattern Recognition (CVPR)}, June 2020.

\bibitem{Menon2020Pulse}
Sachit Menon, Alexandru Damian, Shijia Hu, Nikhil Ravi, and Cynthia Rudin.
\newblock {PULSE}: Self-supervised photo upsampling via latent space
  exploration of generative models.
\newblock In {\em Proceedings of the IEEE/CVF International Conference on
  Computer Vision and Pattern Recognition (CVPR)}, pages 2437--2445, 2020.

\bibitem{richardson2021encoding}
Elad Richardson, Yuval Alaluf, Or Patashnik, Yotam Nitzan, Yaniv Azar, Stav
  Shapiro, and Daniel Cohen-Or.
\newblock {Encoding in Style}: a {StyleGAN} encoder for image-to-image
  translation.
\newblock In {\em Proceedings of the IEEE/CVF International Conference on
  Computer Vision and Pattern Recognition (CVPR)}, June 2021.

\bibitem{Sanyal2021ICCV}
Soubhik Sanyal, Alex Vorobiov, Timo Bolkart, Matthew Loper, Betty Mohler,
  Larry~S. Davis, Javier Romero, and Michael~J. Black.
\newblock Learning realistic human reposing using cyclic self-supervision with
  {3D} shape, pose, and appearance consistency.
\newblock In {\em Proceedings of the IEEE/CVF International Conference on
  Computer Vision (ICCV)}, pages 11138--11147, October 2021.

\bibitem{Schroff2015FaceNet}
Florian Schroff, Dmitry Kalenichenko, and James Philbin.
\newblock {FaceNet}: A unified embedding for face recognition and clustering.
\newblock In {\em Proceedings of the IEEE/CVF International Conference on
  Computer Vision and Pattern Recognition (CVPR)}, pages 815--823, June 2015.

\bibitem{Siarohin2019PAMI}
Aliaksandr Siarohin, Stéphane Lathuilière, Enver Sangineto, and Nicu Sebe.
\newblock Appearance and {Pose-Conditioned} human image generation using
  deformable {GANs}.
\newblock {\em IEEE Transactions on Pattern Analysis and Machine Intelligence},
  2019.

\bibitem{Teterwak2019ICCV}
Piotr Teterwak, Aaron Sarna, Dilip Krishnan, Aaron Maschinot, David Belanger,
  Ce Liu, and William~T. Freeman.
\newblock Boundless: Generative adversarial networks for image extension.
\newblock In {\em Proceedings of the IEEE/CVF International Conference on
  Computer Vision (ICCV)}, October 2019.

\bibitem{tov2021designing}
Omer Tov, Yuval Alaluf, Yotam Nitzan, Or Patashnik, and Daniel Cohen-Or.
\newblock Designing an encoder for stylegan image manipulation.
\newblock {\em ACM Transactions on Graphics (Proceedings of ACM SIGGRAPH)},
  40(4), jul 2021.

\bibitem{Xia2021InversionSurvey}
Weihao Xia, Yulun Zhang, Yujiu Yang, Jing{-}Hao Xue, Bolei Zhou, and
  Ming{-}Hsuan Yang.
\newblock {GAN} inversion: {A} survey.
\newblock {\em CoRR}, abs/2101.05278, 2021.

\bibitem{zhang2018perceptual}
Richard Zhang, Phillip Isola, Alexei~A Efros, Eli Shechtman, and Oliver Wang.
\newblock The unreasonable effectiveness of deep features as a perceptual
  metric.
\newblock In {\em Proceedings of the IEEE/CVF International Conference on
  Computer Vision and Pattern Recognition (CVPR)}, 2018.

\bibitem{Zhao2021CoModGAN}
Shengyu Zhao, Jonathan Cui, Yilun Sheng, Yue Dong, Xiao Liang, Eric~I Chang,
  and Yan Xu.
\newblock Large scale image completion via {Co-Modulated} generative
  adversarial networks.
\newblock In {\em International Conference on Learning Representations (ICLR)},
  2021.

\bibitem{Zhao2021HumanCompletion}
Zibo Zhao, Wen Liu, Yanyu Xu, Xianing Chen, Weixin Luo, Lei Jin, Bohui Zhu,
  Tong Liu, Binqiang Zhao, and Shenghua Gao.
\newblock Prior based human completion.
\newblock In {\em Proceedings of the IEEE/CVF International Conference on
  Computer Vision and Pattern Recognition (CVPR)}, pages 7951--7961, June 2021.

\bibitem{Zhou2021Human}
Qiang Zhou, Shiyin Wang, Yitong Wang, Zilong Huang, and Xinggang Wang.
\newblock Human de-occlusion: Invisible perception and recovery for humans.
\newblock {\em CoRR}, abs/2103.11597, 2021.

\end{thebibliography}


\begin{thebibliography}{10}\itemsep=-1pt

\bibitem{Heusel2017GANs}
Martin Heusel, Hubert Ramsauer, Thomas Unterthiner, Bernhard Nessler, and Sepp
  Hochreiter.
\newblock {GANs} trained by a two time-scale update rule converge to a local
  nash equilibrium.
\newblock {\em Advances in neural information processing systems}, 30, 2017.

\bibitem{Karras2020StyleGAN2}
Tero Karras, Samuli Laine, Miika Aittala, Janne Hellsten, Jaakko Lehtinen, and
  Timo Aila.
\newblock Analyzing and improving the image quality of {StyleGAN}.
\newblock In {\em Proceedings of the IEEE/CVF International Conference on
  Computer Vision and Pattern Recognition (CVPR)}, pages 8107--8116, 2020.

\bibitem{Kynkaanniemi2019PnR}
Tuomas Kynk{\"a}{\"a}nniemi, Tero Karras, Samuli Laine, Jaakko Lehtinen, and
  Timo Aila.
\newblock Improved precision and recall metric for assessing generative models.
\newblock {\em Advances in Neural Information Processing Systems}, 32, 2019.

\bibitem{Liu2016DeepFashion}
Ziwei Liu, Ping Luo, Shi Qiu, Xiaogang Wang, and Xiaoou Tang.
\newblock {DeepFashion}: Powering robust clothes recognition and retrieval with
  rich annotations.
\newblock In {\em Proceedings of the IEEE/CVF International Conference on
  Computer Vision and Pattern Recognition (CVPR)}, June 2016.

\bibitem{Zhao2021CoModGAN}
Shengyu Zhao, Jonathan Cui, Yilun Sheng, Yue Dong, Xiao Liang, Eric~I Chang,
  and Yan Xu.
\newblock Large scale image completion via {Co-Modulated} generative
  adversarial networks.
\newblock In {\em International Conference on Learning Representations (ICLR)},
  2021.

\end{thebibliography}
}

\end{document}


\title{\name for Full-Body Image Generation: Supplementary Materials}

\author{
    \begin{tabular}{c} Anna Fr\"{u}hst\"{u}ck\textsuperscript{{1, 2}} \\ Niloy J. Mitra\textsuperscript{{2, 3}} \end{tabular}
    \begin{tabular}{c} Krishna Kumar Singh\textsuperscript{2} \\ Peter Wonka\textsuperscript{1} \end{tabular}
    \begin{tabular}{c} Eli Shechtman\textsuperscript{2} \\ Jingwan Lu\textsuperscript{2} \end{tabular}\\[2ex]
    \textsuperscript{1~}KAUST \quad 
	\textsuperscript{2~}Adobe Research \quad
	\textsuperscript{3~}University College London\\[1.1ex]
	{\tt\footnotesize anna.fruehstueck@kaust.edu.sa, \{krishsin, elishe\}@adobe.com, \{niloym, pwonka\}@gmail.com, jlu@adobe.com}\\
}

\maketitle

\section{Additional Results and Discussion}
\label{sec:results}
We use a video to illustrate our entire InsetGAN pipeline and to show additional results including a joint latent space walk (see the description below). The video is available at the project webpage \texttt{afruehstueck.github.io/insetgan}. 

\begin{figure}[t]
  \centering
  \includegraphics[width=0.99\linewidth]{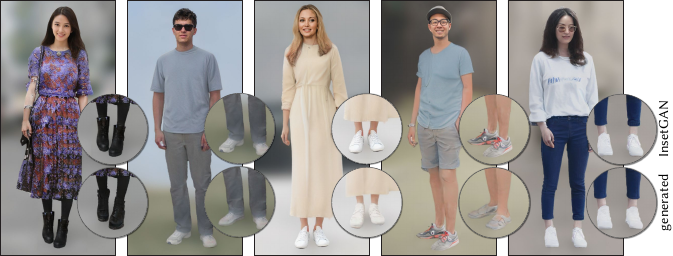}
  \caption{\textbf{Optimizing two Insets.} \name successfully merges a canvas with two distinct insets (shoes and face).
   }
   \label{fig:two_insets}
  \vspace*{-.2in}
\end{figure}

\begin{figure*}[!b]
  \centering
  \includegraphics[width=0.9\linewidth]{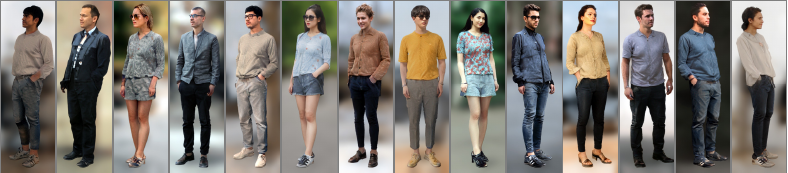}
  \caption{\textbf{Orientations.} We demonstrate that our technique can capture a wide range of face orientations and generate natural-looking face-body compositions that are oriented appropriately for each respective input face.
   }
   \label{fig:face_orientations}
  \vspace*{-.2in}
\end{figure*}

\textbf{Two Insets.}
We demonstrate in \cref{fig:two_insets} that our technique is able to generate good results for insets of another domain beyond faces. To that end, we trained a shoe generator at 256$\times$256px resolution on shoes cropped from the same dataset and use the generated shoe outputs to improve problematic areas in the canvas domain with higher-quality shoe insets exhibiting more detailed and natural features than the full-body GAN. 
These results also demonstrate that our technique can jointly optimize more than one inset: in this example, we select a target face, and target shoes, and find an appropriate body, optimizing all three generators to create a seamless output.

\textbf{Face Orientations.}
While our results exhibit somewhat limited body poses due to the entanglement of body pose variability and deterioration in image quality, we show that we are able to still capture a variety of face orientations and our technique can match the face orientation with a plausible looking body oriented correctly based on the face target, as shown in the results in \cref{fig:face_orientations}.

\begin{figure}[t]
  \centering
   \includegraphics[width=0.98\linewidth]{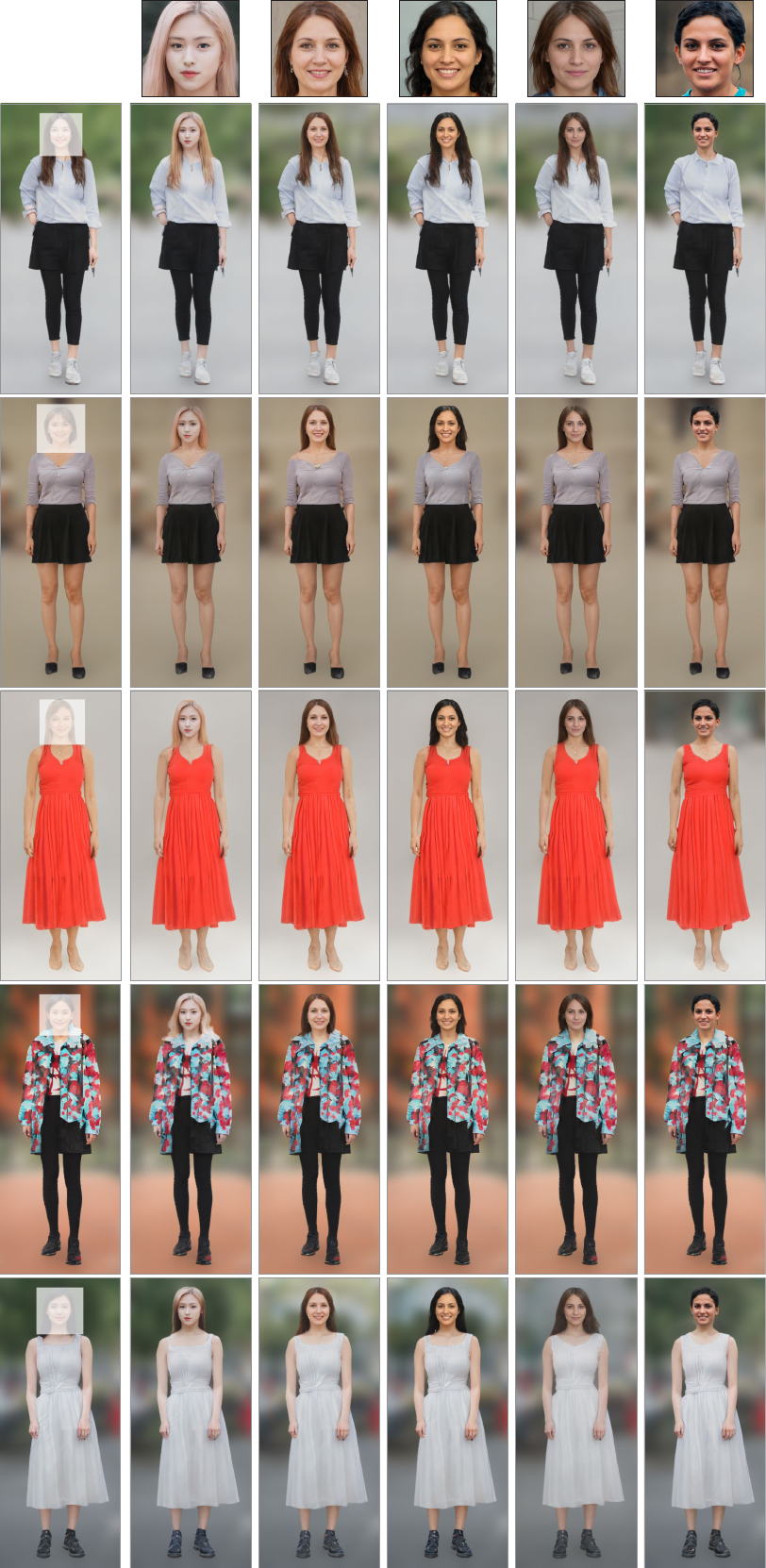}
   \caption{\textbf{Face Body Montage.} Given faces \textit{(top row)} generated by a pretrained FFHQ model and bodies \textit{(left column)} synthesized by our full-body human generator, we apply \textit{joint} latent optimization to find compatible face and human latent codes that are combined to produce coherent full-body humans. 
   }
   \label{fig:face_clothing_matrix2}
   \vspace*{-.25in}
\end{figure}

\begin{figure}[t]
  \centering
   \includegraphics[width=0.98\linewidth]{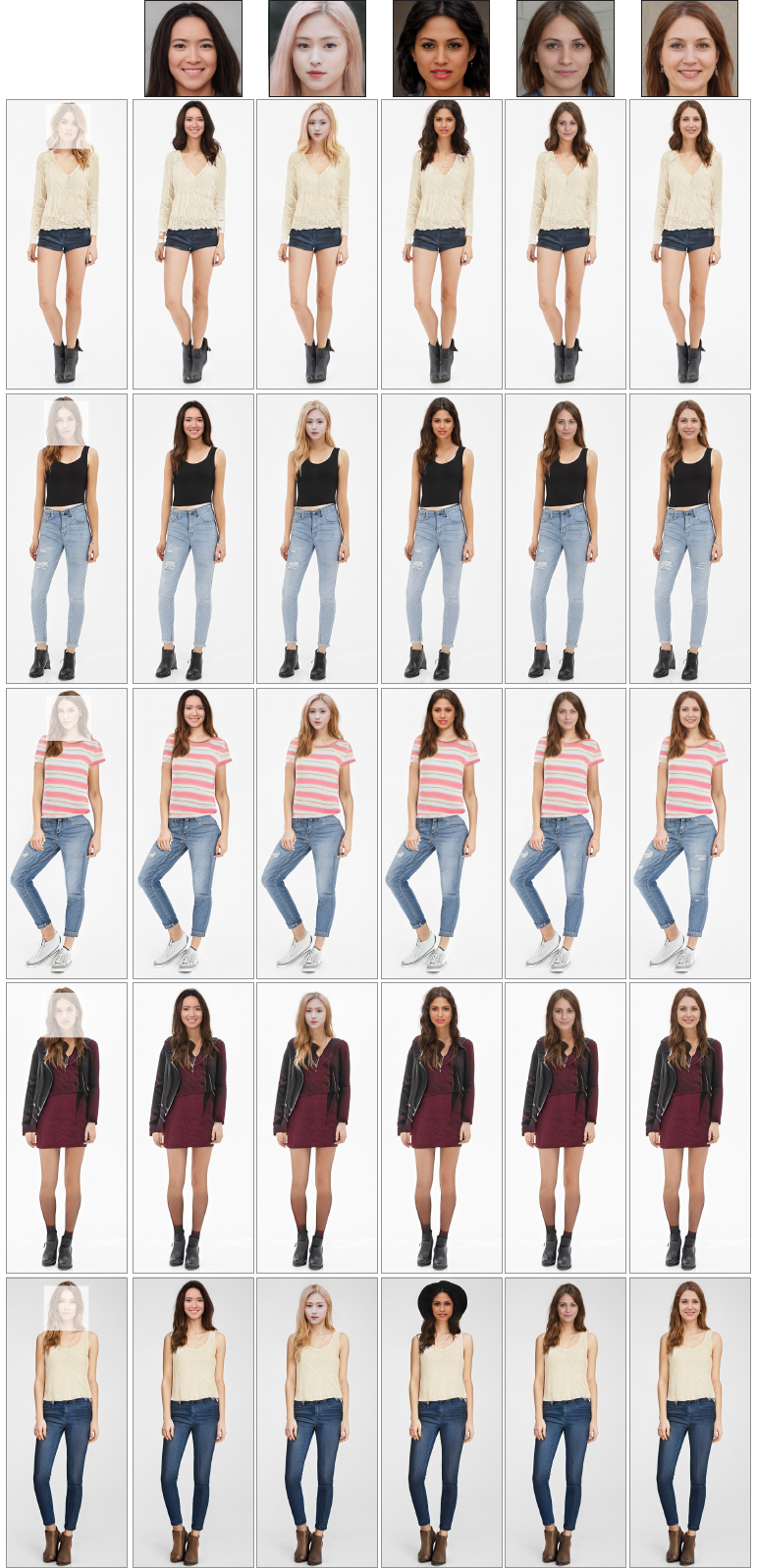}
   \caption{\textbf{Face Body Montage.} Given faces \textit{(top row)} generated by a pretrained FFHQ model and bodies \textit{(left column)} synthesized by full-body DeepFashion~\cite{Liu2016DeepFashion} generator, we apply \textit{joint} latent optimization to find compatible face and human latent codes that are combined to produce coherent full-body humans. 
   }
   \label{fig:face_clothing_matrix1}
   \vspace*{-.25in}
\end{figure}

\textbf{Face-Body Montage.  }
We show some additional results using a human generator trained on our custom dataset as well as another generator trained on the DeepFashion dataset as seen in Figures~\ref{fig:face_clothing_matrix2} and~\ref{fig:face_clothing_matrix1}. The generator trained on Deep Fashion is able to synthesize bodies and garment details with good quality, but the generator is overfitted to the limited quantity and variety of the input data and is thus not very flexible in harmonizing skin tone differences (columns 2 and 3 in \cref{fig:face_clothing_matrix1}).
Note how the results in \cref{fig:face_clothing_matrix2} adapt the hair and body composition, even generating reasonable results for the short-haired woman in the rightmost column.

\newcommand{\kS}{K_{\textit{start}}} 
\newcommand{\kE}{K_{\textit{end}}} 
\newcommand{\kPrev}{K_{\textit{previous}}} 
\newcommand{\wAS}{{\w_{A}}_\textit{start}} 
\newcommand{\wAE}{{\w_{A}}_\textit{end}} 
\newcommand{\wBS}{{\w_{B}}_\textit{start}} 
\newcommand{\wBE}{{\w_{B}}_\textit{end}} 
\newcommand{\wBNext}{{\w_{B}}_\textit{next}} 
\newcommand{\wBPrev}{{\w_{B}}_\textit{previous}} 
\newcommand{\bboxS}{\bbox_\textit{start}} 
\newcommand{\bboxE}{\bbox_\textit{end}} 

\textbf{Latent Space Walk.  }
We are able to create a joint latent space walk (see supplementary video) by linearly interpolating both the face and the body latents and use them as initialization in our joint optimization framework to combine the face and body seamlessly in a temporally coherent way. 
We explain the process to obtain a seamless interpolation in $n$ steps based on two key frames $\kS$ and $\kE$ and their corresponding optimized face and body latent pairs ($\wAS$, $\wBS$) and ($\wAE$, $\wBE$). 
The naive solution is to simply linearly interpolate $n$ times between these latent pairs to obtain the inbetweens. However, interpolating in each of the two latent spaces independently does not yield a seamlessly merged boundary region between the canvas and the inset. In order to improve this boundary, we consider the latent space walk of the canvas as fixed and define the optimization of the inset as an interpolation problem where we optimize frame by frame and do the following at frame $i$: \\

(1) Consider the previous frame (initially, this is $\kS$) and obtain the next frame as the linear interpolation given by $f=\frac{1}{n-i}$ and $\wBNext = (1-f)\times\wBPrev + f\times\wBE$

(2) To avoid unwanted jittering, we no longer reevaluate the face bounding box per frame but linearly interpolate from $\bboxS$ to $\bboxE$ to obtain a smooth transition from one inset position to the next.

(3) Optimize $\wBNext$ for a small number of iterations (e.g. 100 optimization steps) with a set of losses optimizing for (a) the edge coherence with the canvas, (b) the identity preservation with the starting point of $\wBNext$ and (c) the minimization of the edge region changes with respect to the last frame $\kPrev$.

We use this method to insert about 20 to 40 interpolated frames between two given keyframes and render the resulting animation to a video at 16-20fps. By replicating the first keyframe as the last, the latent space walk can loop infinitely.

\begin{figure}[ht]
  \centering
   \includegraphics[width=0.99\linewidth]{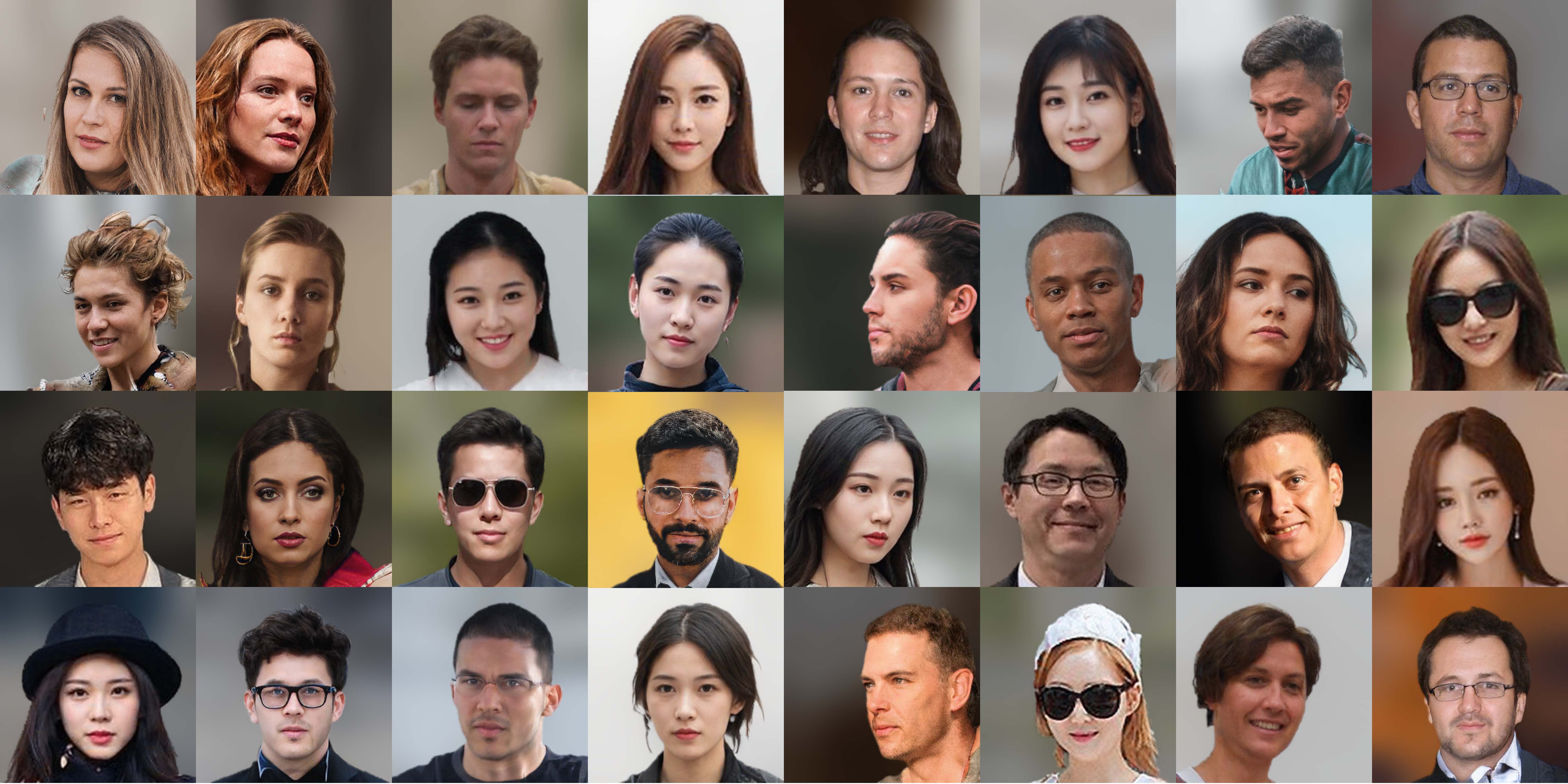}
   \caption{\textbf{Face Synthesis.} We show unconditionally generated results at 256$\times$256px resolution from a face generator trained on the same data as our full-body human generator.
   }
   \label{fig:face_generator}
   \vspace*{-.22in}
\end{figure}

\textbf{Custom Face Generator. }
Most results in the paper and supplementary use a face generator trained on the same data used to train our human GAN. We crop the faces and resample them to 256$\times$256px resolution and train a face generator using the StyleGAN2 architecture. We show some generated face samples in \cref{fig:face_generator}. The visual quality is much higher than that of the faces generated by our full-body generator. Compared to the FFHQ face model, our custom face generator can be used to obtain nicer joint optimization results that better preserve the input face characteristics (ethnicity, skin tone, etc.) without distribution shift. 

\section{Evaluation}

\textbf{Precision and Recall Scores.  }
In addition to calculating the FID scores to quantitatively evaluate our InsetGAN improved results, we also followed Kynkanniemi et al.~\cite{Kynkaanniemi2019PnR} and evaluated the precision and recall score. Precision and Recall provide a more disentangled way of mapping quality and variability of samples. These metrics are a very intuitive quantitative evaluation tool for GANs. Of the two calculated values, precision describes a measurement of image quality (higher=better quality), and recall quantifies the variability of the generated images (high=more variability). Both metrics are scaled between 0 and 1.

We evaluate these scores on more ($t$~=~$0.4$) and less ($t$~=~$0.7$) truncated results to observe the impact of improving overall full-body generation quality at the cost of lowering the variability. All evaluations are performed both on the full-body image as well as on a crop area around the face region that includes a border around the pasted region to evaluate the image coherence. These generated images are compared to the dataset to evaluate a precision and recall score. We calculate the baseline as the precision \& recall of unconditional generation of our model (1), and then evaluate the scores for two different datasets used for improving the face region: (2) the pretrained FFHQ face generator and (3) our custom face dataset trained on the same data as the full-body generator as shown in \cref{fig:face_generator}.

\begin{center}
\vspace*{-.15in}
\small
\begin{tabular}{l c c c c}
\toprule
   $t$ = $0.7$    & \multicolumn{2}{c}{\textbf{Full-body Image}} & \multicolumn{2}{c}{\textbf{Face Crop Area}}\\
                  & Precision       & Recall           & Precision        & Recall           \\
\midrule
(1) unconditional & 0.6958          & \textbf{0.3280}  & 0.7980           &  \textbf{0.2570} \\
(2) FFHQ          & 0.8293          & 0.3126           & 0.8522           &  0.1624          \\
(3) our dataset   & \textbf{0.8364} & 0.3076           & \textbf{0.8891}  &  0.1576          \\
\midrule
   $t$ = $0.4$    & \multicolumn{2}{c}{\textbf{Full-body Image}} & \multicolumn{2}{c}{\textbf{Face Crop Area}}\\
                  & Precision       & Recall           & Precision        & Recall           \\
\midrule
(1) unconditional & 0.9247          & 0.0334           &  0.9206          &  \textbf{0.0552} \\
(2) FFHQ          & \textbf{0.9333} & 0.0336           &  0.9386          &  0.0180          \\
(3) our dataset   & 0.9298          & \textbf{0.0362}  &  \textbf{0.9541} &  0.0182          \\
\bottomrule
\end{tabular}
\end{center}

We can see that our method is able to achieve a significant improvement in precision throughout all experiments. The increase in precision is particularly large for the less-truncated ($t$~=~$0.7$) experiments exhibiting more artifacts in the unconditional generator, where we are able to improve the precision by a large margin using our own model. We can also achieve a comparable improvement using the FFHQ model, which is somewhat surprising since the training data is not based on the same input distribution as the full-body generator. This shows (a) that the generative capabilities of well-trained GANs are providing powerful generalizable models of their domain and (b) that our method is able to encourage good results even for specialized part generators that are trained on a completely different distribution.
Note that our improvements come at a cost of a small drop in recall, which denotes that the variability of the samples goes down a little in most instances. The drop is insignificant when measured on the full body, yet noticeable when calculating the metrics on the cropped face area. We attribute this drop in recall to the fact that we reduce artifacts caused by outliers and unusual generated samples in the images, which decreases the variability in the samples. 
Our results on larger truncation ($t$ = $0.4$) paint a similar picture, albeit with smaller margins in the improvement, as the generator is significantly more restricted at this setting. We can also see that the Recall values at this truncation level are already extremely low.

\textbf{Issues of FID and Training Data.  }
As shown in the main paper, the FID score~\cite{Heusel2017GANs} of unconditionally generated samples is similar to that of InsetGAN improved samples. We think there are two main reasons: (1) Even though FID is the most commonly used metric for evaluating image generation quality, it does not correlate well with human perceptual quality and cannot effectively capture subtle visual differences as discussed in \cite{Zhao2021CoModGAN}. Our user study results also contradict results based on FID because in the user study our InsetGAN results are clearly preferred by the users but FID cannot properly reflect the quality improvements; (2)

\begin{figure}[t]
  \centering
   \includegraphics[width=0.99\linewidth]{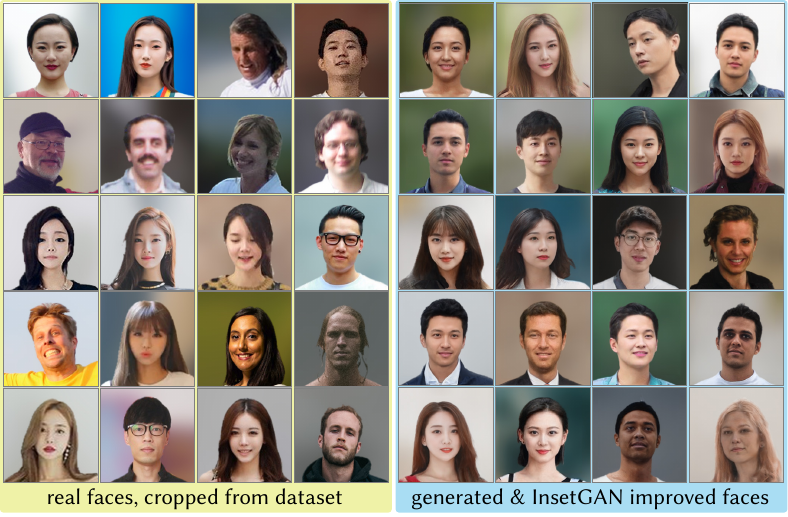}
   \caption{\textbf{Dataset Quality. } We show a comparison of faces cropped from our dataset \textit{(left)} with faces sampled from unconditionally generated and InsetGAN-improved humans \textit{(right)}. Zoom in to observe the variable quality of the input data.
   }
   \label{fig:dataset_quality}
   \vspace{-.2in}
\end{figure}

Our dataset contains photographs of varying quality and resolution, ranging from high-resolution studio quality photographs to low-lighting cellphone snapshots. Additionally, human subjects might only occupy small regions of the original photographs. After cropping and resampling, artifacts can be quite visible and sometimes magnified. For instance, as shown in \cref{fig:dataset_quality} left, we observe JPG artifacts, motion blur and noisiness caused by low-lighting condition. Our face GAN trained on cropped face regions from this dataset can alleviate some of these artifacts when used to improve the generated faces from the trained human GAN as shown in \cref{fig:dataset_quality} right. We notice a good number of our randomly-sampled 4K training images used for FID evaluation contain artifacts. The human GAN generated faces that contain more artifacts might accidentally have more similar distribution to the training set than the nice clean faces generated by the face GAN used in our joint optimization. 

\begin{figure}[t]
  \centering
   \includegraphics[width=0.99\linewidth]{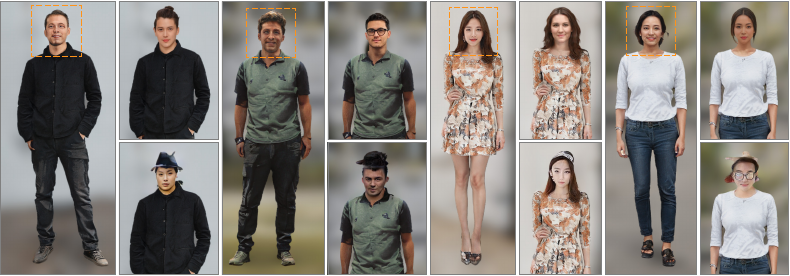}
   \caption{\textbf{Improved Comparison with CoModGAN. } We remove the conditional constraint on the face and allow for unconditional (only edge-conditional) face insertion. In contrast to Fig. 9 of the main paper, we see that the face is allowed to diverge from the face input. We also keep the body latent fixed so that the body pixels in both our results and \comodgan results remain unchanged.
   }
   \label{fig:comodgan_face_comparison}
   \vspace{-.2in}
\end{figure}

\textbf{Additional Comparison with \comodgan.  } 
In \cref{fig:comodgan_face_comparison}, we show an additional comparison of our method to \comodgan where we evaluate the quality of our "inpainting" capabilities after removing the constraint that the output face needs to be similar to the underlying input face. We only optimize the face latent code based on the edge coherence term and keep the body latent code fixed. This makes the comparison fairer, since \comodgan invents completely new faces based only on the context pixels outside the input bounding box and does not alter the pixels of the body. We show that we are able to generate plausible and coherent results without using the input face as guidance and without joint optimization.

\textbf{User Study Details.  }
We provide additional details about the user studies we conducted on Amazon Mechanical Turk. We adopt a forced choice paired comparison procedure where the participant is shown a pair of images at a time and is asked to ``select in which of the two images the person looks more plausible and real" as seen in \cref{fig:userstudy_interface}. For each HIT (Human Intelligence Task), we randomize the pair order and whether our result is on the left or right.

We performed four different independent studies:\\

(1) Compare unconditionally generated samples (truncated with $t$ = $0.4$) with images in our training set. 

(2) Compare unconditionally generated samples ($t$ = $0.4$) with the results of joint \name optimization for face refinement.

(3) Compare unconditionally generated samples ($t$ = $0.4$) with the results of using \comodgan for face regeneration.

(4) Compare our InsetGAN results with \comodgan results directly.

In study (1) the images are unpaired since there is no correspondence between any generated image and any training image. Given two images, we show the first one for a second and then the other one for another second. The images are rendered at 384$\times$768px resolution so that they fit into the browser without the need of scrolling.\\
For studies (2), (3) and (4), we show 512$\times$1024px center-cropped images side-by-side to the participants so that they can focus on the differences of the image details. The selection buttons above the image pairs are faded in after a 6-second delay, so that users are encouraged to carefully study the image differences before making their selections. We collect 5 votes per image pair and choose the winner image that receives 3 or more votes. We summarize the results into the following table:

\begin{figure}[t]
  \centering
   \includegraphics[width=0.7\linewidth]{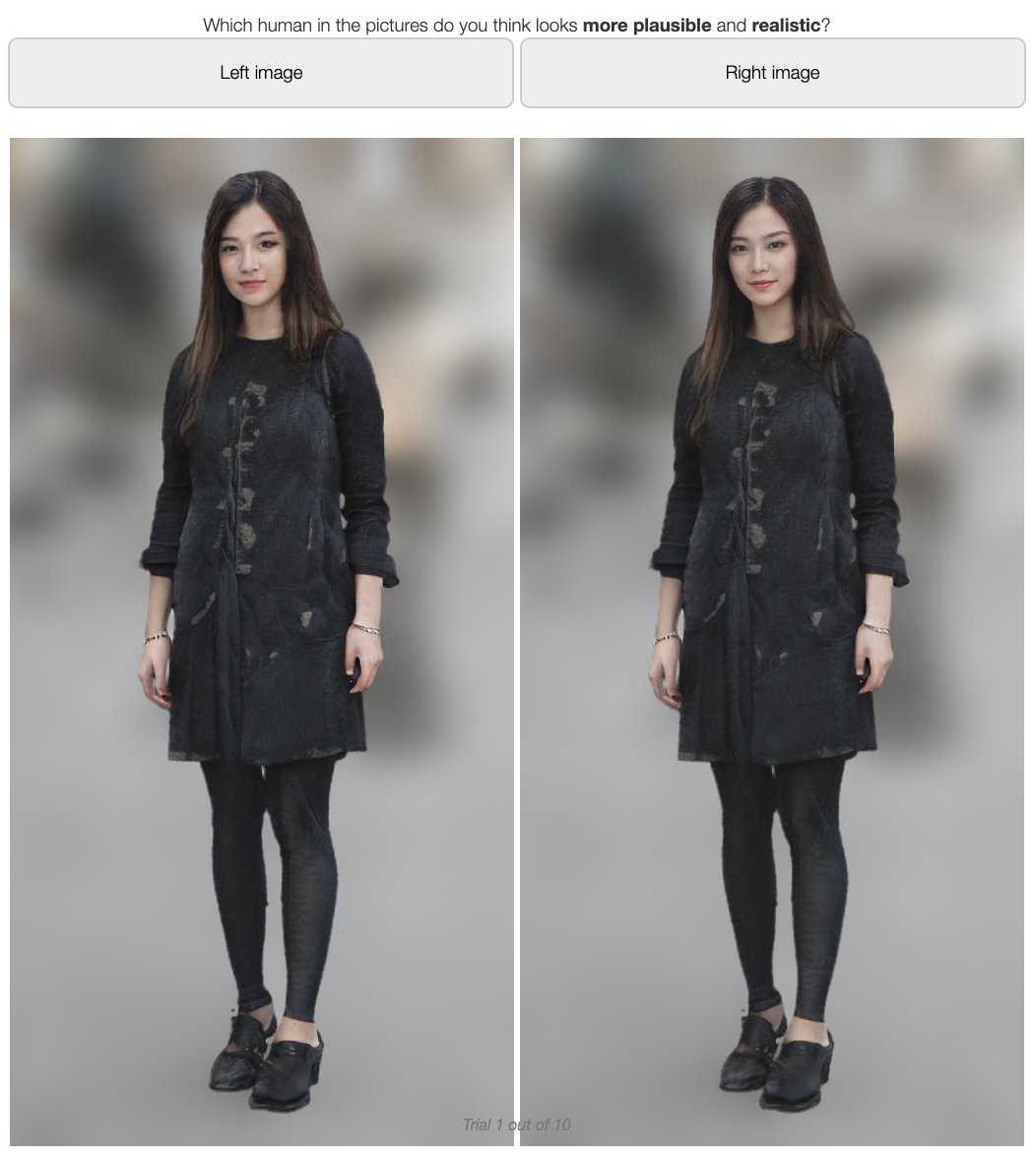}
   \caption{\textbf{User Study Interface.} We show the web interface presented to participants of our user study for task (2).
   }
   \label{fig:userstudy_interface}
   \vspace{-.1in}
\end{figure}

\begin{center}
\footnotesize
\begin{tabular}{*5r}
\toprule
{Study} &  \multicolumn{2}{c}{A} & \multicolumn{2}{c}{B}\\
\midrule
Real (A) / Generated (B)     & 438 & 87.6\% & 62  & 12.4\%  \\
Generated (A) / \name (B) & 10 & 2.0\%  & 490 & 98.0\% \\
Generated (A) / \comodgan (B) & 465 & 93.0\%  & 35 & 7.0\%   \\
\name (A) / \comodgan (B)  & 495 & 99.0\%  & 5 & 1.0\%   \\
\bottomrule
\end{tabular}
\end{center}

\section{Implementation Details}

\textbf{Unconditional Generation and Adaptive Truncation.  }

\begin{figure}[t]
  \centering
   \includegraphics[width=0.99\linewidth]{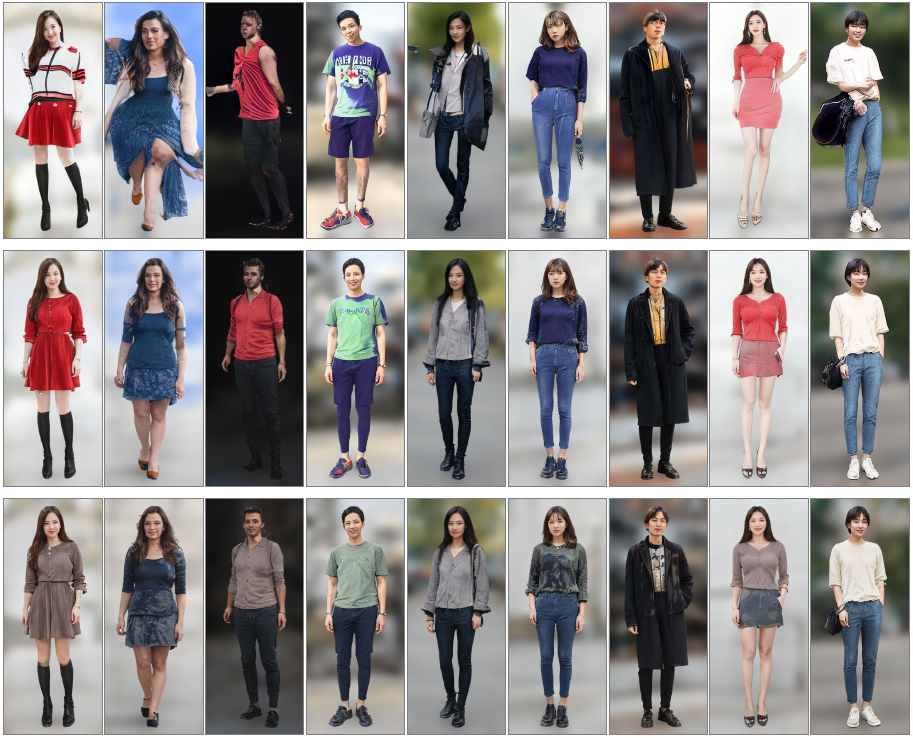}
   \caption{\textbf{Adaptive Truncation. } We show a set of untruncated samples from our human generator exhibiting unrealistic poses and unwanted artifacts. Standard truncation ($t$=$0.6$, \textit{bottom row}) reduces artifacts, but also removes desirable clothing details and reduces the color vibrancy. Our adaptive truncation \textit{(center row)} better preserves colors, texture details and accessories.
   }
   \label{fig:adaptive_truncation}
   \vspace{-.2in}
\end{figure}

Since our generator is trained on very diverse data, we can observe a wide range of image quality when generating untruncated output. 
When truncating the generated results as described in the original StyleGAN2 paper~\cite{Karras2020StyleGAN2} by linearly interpolating from the sample position in $\vec{w}$ space to the average latent $\wAvg$ we can drastically reduce artifacts in pose and details. However, this trick also reduces the diversity in the sample output, and notably reduces the color vibrancy of the output images, as outfit colors are interpolated towards an averaged greyish hue.
In our approach, whenever possible (i.e. whenever we are not constrained to operate in the $\vec{w}$ space), we use a layer-adaptive truncation scheme to generate visually pleasing result of improved perceptual quality while preserving as many diverse features as possible from the untruncated samples, as shown in \cref{fig:adaptive_truncation}.

To achieve this, when generating unconditional samples, we use the $\vec{w}^{+}$ space and define a separate truncation value for each layer. In our generator, we have 18 layers, and we define the layer-wise truncation values as  
{
\small
\setlength{\abovedisplayskip}{6pt}
\setlength{\belowdisplayskip}{\abovedisplayskip}
\setlength{\abovedisplayshortskip}{0pt}
\setlength{\belowdisplayshortskip}{3pt}
\begin{align*}
t = [ &0.35, 0.25, 0.25, 0.70, 0.75, 0.65, 0.65, 0.40, 0.40, \\
      &0.35, 0.25, 0.15, 0.15, 0.05, 0.05, 0.05, 0.05, 0.05 ]
\end{align*}
}%
The values were chosen through experimentation where we truncate individual layers separately to identify the ones that cause the most artifacts.
Note that we apply almost no truncation on later layers, as they can be used to generate desirable clothing details, vibrant colors and accessories and do not cause significant artifacts. We observe in our experiments that latent codes for the middle layers (4-7) of the network are most responsible for artifacts, so we truncate them the most.

We also measure the Fréchet Inception Distance (FID) of 4K random results generated using our adaptive truncation scheme and observe a significantly lower FID ($53.26$) as compared to using regular truncation at $t$=$0.6$ ($71.89$). 
We would like to point out that we did not use the adaptive truncation trick when we performed the quantitative evaluations in the main paper, both for clarity and simplicity and because we were optimizing in $\vec{w}+\delta_i$ space, which restricts the effect of adaptive truncation.

\textbf{Optimization details. } 
All our results were optimized using ADAM. We usually stop the optimization when the edge loss falls below a certain threshold (usually defined as $\Lone(\textit{border})(\wB) < 0.09$) or after the number of iterations exceed a threshold (typically 1000 optimization steps).
When performing joint optimization, we define two distinct optimizers for $\wA$ and $\wB$ and switch the optimization target every 50 iterations. Depending on the application, we can start with the canvas optimizer or the inset optimizer.
We choose different learning rates for the canvas optimizer: $lr = 0.05$ and for the inset optimizer: $lr = 0.002$.
We reevaluate the bounding box every 25 iterations during optimization for a certain number of iterations (typically 150 iterations during body generation, 75 iterations during face refinement.) before keeping the bounding box fixed. We observe that reevaluating the bounding box too often or too long makes the optimization unstable.

\textbf{Lambda weights for Losses.  } 
We report the $\lambda$ weight combination we use for the face body montage application. In this use case, we have losses for improving the coherence of $\gA$ and $\gB$ from the perspective of each GAN, as well as losses for controlling the appearance of each output, either by constraining closeness of the center image region (face) to a target or some outer image region (body) to adhere to a specific body. 

\begin{center}
\small
\begin{tabular}{rrcc}
\toprule
{$\lambda$}                & Loss Description & $L_{\textit{body}}$   & $L_{\textit{face}}$\\
\midrule
$\lambda_1$ & $\Lone$                          &  500                & 500               \\
$\lambda_2$ & $\percep$                        &  0.05               & 0.05              \\
$\lambda_3$ & $\percep(\textit{border})$       &  -                  & 0.1               \\
$\lambda_4$ & $\Lone(\textit{border})$         &  2500               & 10000             \\
$\lambda_{r1}$ & $\lR$                         &  25000              & -                 \\
$\lambda_5$ & $\Lone(\textit{target\_body})$   &  9000               & -                 \\
$\lambda_6$ & $\percep(\textit{target\_body})$ &  0.1                & -                 \\
$\lambda_7$ & $\Lone(\textit{target\_face})$   &  -                  & 5000              \\
$\lambda_8$ & $\percep(\textit{target\_face})$ &  -                  & 1.75              \\
\bottomrule
\end{tabular}
\end{center}

We define several different optimization targets that require custom parameters and setup:
\begin{enumerate}
    \item \textbf{Improving the face area of a given human image.}\\
    We either run one-way optimization or take a small learning rate for the human optimizer to allow for a small wiggling of the canvas area around the inset, which generally improves the coherence of inset and canvas. We start with optimizing the inset from a random starting point for a certain number of iterations (e.g. $n$=$100$) and then fix the inner face area by adding an additional loss constraint keeping the face close to the remembered state. This allows more iterations for the boundary area to improve but prevents overfitting to unwanted artifacts in the face region of the input canvas or deterioration of the facial quality due to over-optimization.
    \item \textbf{Finding suitable bodies for a given input face.}\\
    We start with optimizing the canvas from a random starting point for a certain amount of iterations (e.g. $n$=$150$), allowing the body generator to roughly hallucinate a person with similar facial structure as the target. Then, we switch to an alternating optimization schedule, allowing the appearance of body and face to gradually resemble each other in the boundary regions. We can regularize the appearance of the body using a similar strategy as described above to avoid over-optimization.
    \item \textbf{Seamlessly combine a given face and body}
    In order to maintain the appearance of both the input face and the body, we constrain the joint optimization so that the face GAN result stays close to the input face and the body GAN result outside of the face stays close to the input body. 
\end{enumerate}

{
\small
\bibliographystyle{ieee_fullname}
\bibliography{human_synthesis_supp}
}